\documentclass{article}

\usepackage[final]{neurips_2022}

\usepackage[utf8]{inputenc} 
\usepackage[T1]{fontenc}    
\usepackage{hyperref}       
\usepackage{url}            
\usepackage{booktabs}       
\usepackage{amsfonts}       
\usepackage{nicefrac}       
\usepackage{microtype}      
\usepackage{xcolor}         
\usepackage{mathtools}
\usepackage{dsfont}
\usepackage{wrapfig}
\usepackage{multirow}
\usepackage{algorithm}
\usepackage{algorithmicx}
\usepackage{algpseudocodex}

\newcommand{\one}{\mathds{1}}
\newcommand{\res}[2]{$#1$ \scalebox{0.75}{$\,\pm\,#2$}}
\newcommand{\bres}[2]{$\mathbf{#1}$ \scalebox{0.78}{$\,\pm\,#2$}}
\setcitestyle{numbers,square}

\definecolor{gray}{rgb}{0.5,0.5,0.5}
\newcommand{\minihead}[1]{{\noindent\textbf{#1.} }}
\newcommand\ExtraSep{\dimexpr\cmidrulewidth+\aboverulesep+\belowrulesep\relax}

\newcommand\blfootnote[1]{%
  \begin{NoHyper}%
  \renewcommand\thefootnote{}\footnote{#1}%
  \addtocounter{footnote}{-1}%
  \end{NoHyper}%
}

\title{Spartan: Differentiable Sparsity \\ via Regularized Transportation}
\author{
  Kai Sheng Tai\\
  Meta AI \\
  \And
  Taipeng Tian \\
  Meta AI \\
  \And
  Ser-Nam Lim \\
  Meta AI \\
}

\begin{document}
\maketitle

\begin{abstract}
    We present Spartan, a method for training sparse neural network models with a predetermined level of sparsity.
    Spartan is based on a combination of two techniques: (1) soft top-$k$ masking of low-magnitude parameters via a regularized optimal transportation problem and 
    (2) dual averaging-based parameter updates with hard sparsification in the forward pass.
    This scheme realizes an exploration-exploitation tradeoff: early in training, the learner is able to explore various
    sparsity patterns, and as the soft top-$k$ approximation is gradually sharpened over the course of training, the balance
    shifts towards parameter optimization with respect to a fixed sparsity mask.
    Spartan is sufficiently flexible to accommodate a variety of sparsity allocation policies, 
    including both unstructured and block structured sparsity, 
    as well as general cost-sensitive sparsity allocation mediated by linear models of per-parameter costs.
    On ImageNet-1K classification, Spartan yields 95\% sparse ResNet-50 models and 90\% block sparse ViT-B/16 models
    while incurring absolute top-1 accuracy losses of less than 1\% compared to fully dense training.\blfootnote{Correspondence to: Kai Sheng Tai\hspace{0.5em}(\texttt{kst@meta.com})} 
\end{abstract}

\section{Introduction}

Sparse learning algorithms search for model parameters that minimize training loss while retaining only a small fraction
 of non-zero entries.
 Parameter sparsity yields benefits along several axes: reduced model storage costs, greater computational and energy efficiency 
 during training and inference, and potentially improved model generalization~\citep{mozer1988skeletonization}.
However, sparse training is a challenging optimization problem: for the general problem of learning parameters $\theta\in\mathbb{R}^d$ subject to the constraint that $\theta$ is $k$-sparse,
the feasible set is the union of $\binom{d}{k}$ axis-aligned hyperplanes intersecting at the origin, each of dimension $k$.
This complex, nonconvex geometry of the feasible set compounds the existing difficulty of optimizing deep neural network models.

This work focuses on improving the generalization error of sparse neural network models.
To this end, we introduce \emph{Spartan}, or \emph{Sparsity via Regularized Transportation}---a sparse learning algorithm that
leverages an optimal transportation-based top-$k$ masking operation to induce parameter sparsity during training.
Spartan belongs to the family of ``parameter dense'' training algorithms that maintains a dense parameter vector $\theta \in \mathbb{R}^d$
throughout training~\citep{zhu2018prune,jayakumar2020top}, in contrast to ``parameter sparse'' training algorithms that adhere to a $\tilde{\mathcal{O}}(k)$
memory budget for representing the parameters of a $k$-sparse model~\citep{bellec2018deep,mocanu2018scalable,mostafa2019parameter,dettmers2019sparse,evci2020rigging}.
While computational cost and memory usage at training time are important considerations, Spartan primarily optimizes for performance at inference time.

Intuitively, Spartan aims to achieve a controlled transition between the \emph{exploration} and \emph{exploitation} of various sparsity patterns during training.
In the exploration regime, the goal is for the learner to easily transition between differing sparsity patterns in order to 
escape those that correspond to poor minima of the loss function.
On the other hand, the goal of exploitation is for the learner to optimize model parameters given a fixed sparsity pattern, or a small set of similar patterns.
This latter setting avoids frequent oscillations between disparate sparsity patterns, which may be detrimental to the optimization process.

\begin{figure}
    \centering
    $\vcenter{\hbox{\includegraphics[width=0.65\textwidth]{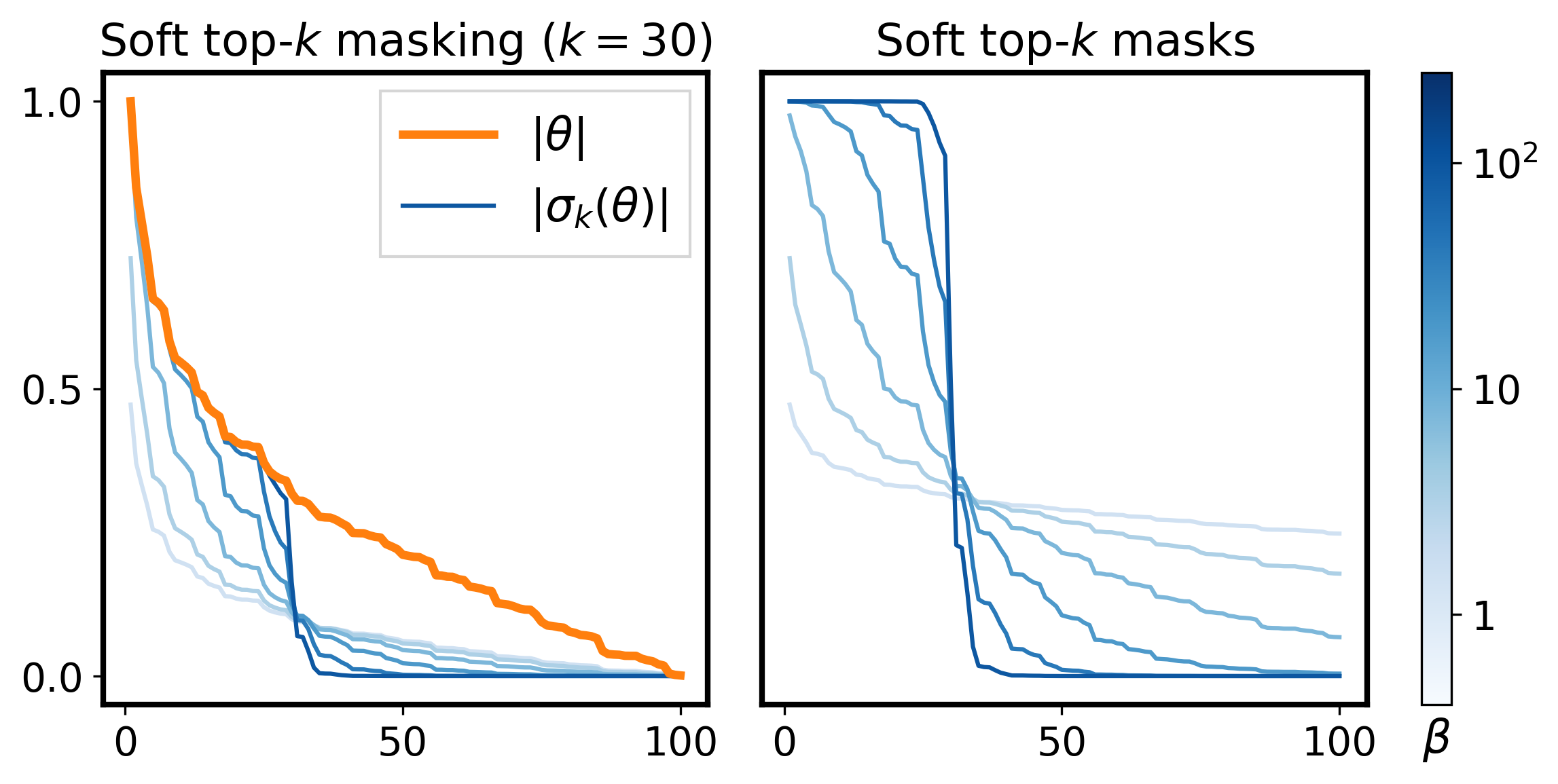}}}$
    \hspace*{0.1em}
    $\vcenter{\hbox{\includegraphics[width=0.33\textwidth]{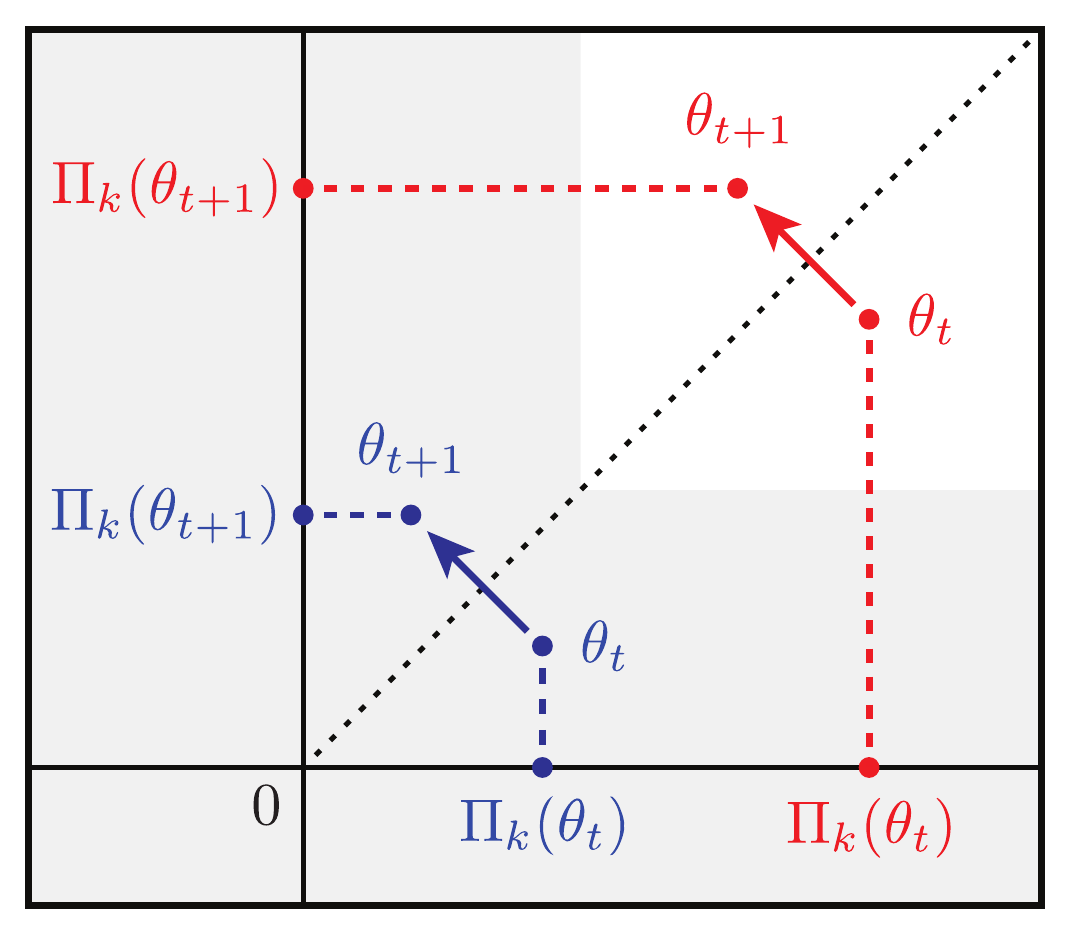}}}$
    \hspace{0.2in}(a)\hspace{1.35in}(b)\hspace{1.9in}(c)
    \caption{\textbf{Soft top-$k$ masking.} {(a, b)}~The soft top-$k$ masking operator $\sigma_k^\beta$ computes approximately $k$-sparse outputs, with the sharpness of the mask controlled by the parameter $\beta$.
    {(c)}~Small updates to iterates far from the $k$-sparse feasible set ({\color{red}$\theta_t \rightarrow \theta_{t+1}$}) can correspond to a large perturbation in parameter space after projection by $\Pi_k$.
    Updates to iterates {\color{blue}$\theta_t$} in the approximately sparse region (shaded in grey) correspond to smaller post-projection perturbations.
    }
    \label{fig:intro}
\end{figure}

\begin{figure}
    \centering
    \includegraphics[width=0.99\linewidth]{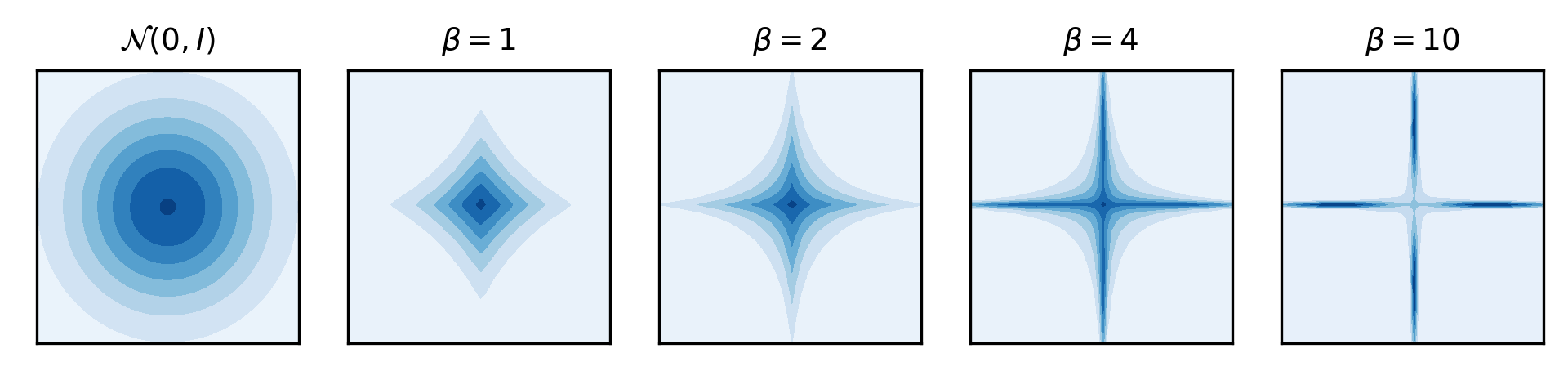}
    \vspace{-0.5em}
    \caption{\textbf{A 2D example of soft masking.}
    We plot probability densities corresponding to the action of the soft top-1 operator $\sigma_1^\beta$ on the 2D standard Gaussian distribution (darker colors indicate higher density).
    Specifically, we visualize the densities of $\sigma^\beta_1(x) \mid x \sim \mathcal{N}(0, I_2)$ for a range of sharpness parameters $\beta$.
    Higher values of $\beta$ constrain iterates to be closer to the $k$-sparse feasible set.}
    \label{fig:gaussian}
\end{figure}

We operationalize this intuition through the use of a differentiable \emph{soft top-$k$ masking} operation $\sigma_{k}^\beta : \mathbb{R}^d \rightarrow \mathbb{R}^d$.
This function maps parameters $\theta$ to an approximately sparse output that suppresses low-magnitude entries in $\theta$.
We parameterize the soft top-$k$ mask with a \emph{sharpness parameter} $\beta$: at $\beta = 0$, $\sigma_k^{\beta}$ simply scales the input by a constant,
and as $\beta \rightarrow \infty$, the mapping reduces to hard top-$k$ magnitude-based selection~(Figure~\ref{fig:intro}~(a, b)).
Soft top-$k$ masking therefore constrains the iterates $\theta_t$ to be close to the set of exactly $k$-sparse vectors, with the strength of this constraint mediated by the parameter $\beta$.
Figures~\ref{fig:intro}~(c)~and~\ref{fig:gaussian} give some geometeric intuition for the effect of this mapping.
We implement soft top-$k$ masking using a regularized optimal transportation formulation~\citep{cuturi2013sinkhorn,xie2020differentiable}
and demonstrate that this technique scales to networks on the order of $10^8$ parameters in size.

We evaluate Spartan on ResNet-50~\citep{he2016deep} and ViT~\citep{dosovitskiy2020image} models trained on the ImageNet-1K dataset.
On ResNet-50, we find that sparse models trained with Spartan achieve higher generalization accuracies than those trained with existing
methods at sparsity levels of 90\% and above.
In particular, we train ResNet-50 models to $76.5\%$ top-1 validation accuracy at 95\% sparsity and to $74.2\%$ accuracy at 97.5\% sparsity, improving on the previous
state-of-the-art by $0.6\%$ and $1.6\%$ respectively.
Our sparse ViT-B/16 models reduce model storage size by $10\times$ and inference FLOPs by $7.4\times$ at the cost of a $0.6\%$ accuracy reduction relative to DeiT-B~\citep{touvron2021training}.
We further demonstrate that Spartan is effective for block structured pruning, a form of structured sparsity that is more amenable to
acceleration on current GPU hardware than unstructured sparsity~\citep{gray2017gpu}. On a ViT-B/16 model with $16 \times 16$ block structured pruning, Spartan achieves
$79.1\%$ top-1 accuracy at 90\% sparsity, compared to the baseline of $74.1\%$ at the same sparsity level.

To summarize, we make the following contributions in this paper:
\begin{itemize}
    \item We present Spartan, a sparsity-inducing training algorithm based on a soft top-$k$ masking operation. We show that Spartan interpolates between two existing sparse learning algorithms: iterative magnitude pruning~\citep{zhu2018prune} and Top-$K$ Always Sparse Training~\citep{jayakumar2020top}.
    \item \looseness=-1 We empirically evaluate Spartan using ResNet-50 and ViT models trained on the ImageNet-1K dataset, demonstrating consistent improvements over the prior state-of-the-art.
    \item We study the effect of Spartan's hyperparameters on an exploration-exploitation tradeoff during training and on the final accuracy of the trained models.
\end{itemize}

We provide an open source implementation of Spartan at \url{https://github.com/facebookresearch/spartan}.

\minihead{Notation} We use $\mathbb{R}_+$ and $\mathbb{R}_{++}$ to denote the nonnegative real numbers and the strictly positive real numbers respectively. 
$\one_d$ is the all-ones vector of dimension $d$.
Given a vector $x\in\mathbb{R}^d$, $\|x\|_p$ denotes the $p$-norm of $x$, $\|x\|_0$ is the number of nonzero entries in $x$, and $|x|$ is the vector of elementwise absolute values. 
When $x$ and $y$ are vectors, $x \odot y$ denotes elementwise multiplication and $x / y$ denotes elementwise division.
The operator $\Pi_k(x) \coloneqq \arg\min_{y : \|y\|_0 \leq k} \|x - y\|_2$ denotes Euclidean projection onto the set of $k$-sparse vectors.

\section{Related Work}

\minihead{Neural network pruning} Our use of a magnitude-based criterion for sparse learning draws on early work in the area of 
neural network pruning~\citep{janowsky1989pruning}. Magnitude-based pruning is computationally cheap relative to alternative
criteria that rely on first- or second-order information~\citep{mozer1988skeletonization,lecun1989optimal,hassibi1992second}, and is a perhaps surprisingly performant
option despite its simplicity~\citep{gale2019state}. More generally, Spartan builds on a substantial body of previous work that aims to jointly optimize
sparsity patterns and model parameters for deep neural networks~[\citealp{yu2012exploiting,collins2014memory,han2015learning,louizos2018learning,lee2018snip,zhu2018prune}, \emph{inter alia}; see, \emph{e.g.},~\citealp{hoefler2021sparsity} for a survey].

\minihead{Spartan as a generalization of existing methods} We highlight two particularly relevant sparse training methods in the literature:
the \emph{iterative magnitude pruning} (IMP) method (Algorithm~\ref{alg:imp},~\citep{zhu2018prune}), and \emph{Top-$K$ Always Sparse Training}, or Top-KAST (Algorithm~\ref{alg:da},~\citep{jayakumar2020top}):
\vspace{-0.5em}
\begin{figure}[H]
    \begin{minipage}{0.49\textwidth}
        \begin{algorithm}[H]
            \caption{Iterative magnitude pruning update}
            \label{alg:imp}
            \begin{algorithmic}[1]
                \Require parameters $\theta_t$, loss function $L(\theta)$, sparsity budget $k$, step size $\eta_t$
                \Ensure parameters $\theta_{t+1}$
                \State $\tilde{\theta}_t = \Pi_k\left(\theta_t\right)$ 
                \State $\theta_{t+1} = \theta_t - \eta_t \nabla \tilde{\theta}_t \nabla L(\tilde{\theta}_t)$
                \State \textbf{return} $\theta_{t+1}$
            \end{algorithmic}
        \end{algorithm}
    \end{minipage}
    \hfill
    \begin{minipage}{0.49\textwidth}
        \begin{algorithm}[H]
            \caption{Dual averaging / Top-KAST update}
            \label{alg:da}
            \begin{algorithmic}[1]
                \Require parameters $\theta_t$, loss function $L(\theta)$, sparsity budget $k$, step size $\eta_t$
                \Ensure parameters $\theta_{t+1}$
                \State $\tilde{\theta}_t = \Pi_k\left(\theta_t\right)$ 
                \State $\theta_{t+1} = \theta_t - \eta_t \nabla L(\tilde{\theta}_t)$ 
                \State \textbf{return} $\theta_{t+1}$
            \end{algorithmic}
        \end{algorithm}
    \end{minipage}
\end{figure}
\vspace{-0.5em}
Note that both Algorithms~\ref{alg:imp} and \ref{alg:da} compute the forward pass through the network using the sparsified parameters $\tilde{\theta}$, obtained by
setting all but the top-$k$ entries of $\theta$ by magnitude to zero. Algorithms~\ref{alg:imp} and \ref{alg:da} differ in the presence of the $\nabla \tilde{\theta}_t$ term in line 2.
This is a $d\times d$ diagonal matrix where $(\nabla \tilde{\theta}_t)_{ii} = 1$ iff $\theta_{t,i}$ was not set to zero by the projection $\Pi_k$, and $(\nabla \tilde{\theta}_t)_{ii} = 0$ otherwise.

At the extreme points of the sharpness parameter $\beta$, Spartan's parameter update procedure reduces to those of IMP and Top-KAST. 
We can thus view Spartan as a method that generalizes and smoothly interpolates between this pair of algorithms. As $\beta \rightarrow \infty$, Spartan is equivalent to IMP:
this approach sparsifies parameters using a top-$k$ magnitude-based binary mask, and entries that were masked in the forward pass receive zero gradient in the backward pass.
At $\beta = 0$, Spartan reduces to Top-KAST: 
this method again sparsifies parameters by magnitude with a binary mask, but unlike IMP, all entries are updated in the backward pass using the gradient of the loss with respect to the \emph{masked} parameters.
The Top-KAST update is thus an application of the \emph{straight-through gradient method}~\citep{bengio2013estimating,courbariaux2015binaryconnect},
otherwise known as \emph{lazy projection} or \emph{dual averaging} (DA) in optimization~\citep{nesterov2009primal,xiao2009dual,bai2018proxquant,dockhorn2021demystifying}. See Sec.~\ref{sec:discussion} for further discussion on this point.

\minihead{Smooth approximations}
Our use of the soft top-$k$ operation is related to prior methods that use the logistic sigmoid function as a differentiable approximation
to the step function for sparse training~\citep{luo2020autopruner,savarese2020winning,azarian2020learned}.
These approaches similarly regulate the sharpness of the approximation with a temperature parameter that scales the input logits to the sigmoid function.
A distinctive characteristic of Spartan is that the parameter $k$ directly controls the degree of sparsity of the mask; 
this differs from previous approaches involving the logistic sigmoid approximation that only indirectly control the degree of sparsity using an $L_1$ penalty term.

\minihead{Transformer-specific methods}
Several approaches are specialized to transformer-based architectures.
These include structured pruning of pre-trained transformers for NLP tasks~\citep{wang2020structured,lagunas2021block,xia2022structured}
and sparse training methods specifically applied to vision transformers~\citep{chen2021chasing,chavan2022vision}. 
In contrast, Spartan is a general-purpose sparse training algorithm that is designed to be agnostic to the model architecture.

\section{Sparsity via Regularized Transportation}
\label{sec:method}

We begin by motivating our proposed approach. A key advantage of the dual averaging method (Algorithm~\ref{alg:da}) over IMP (Algorithm~\ref{alg:imp}) is that it mitigates the issue of gradient sparsity. 
In the IMP update, only those parameters that were not masked in the forward pass receive a nonzero gradient, resulting in slow progress at high levels of sparsity.
In contrast, dual averaging applies a dense update to the parameter vector $\theta$ in each iteration---intuitively, these dense updates can help ``revive'' parameters
that are below the top-$k$ magnitude threshold and are consequently masked to zero.
Dual averaging iterates are thus able to more quickly explore the space of possible sparsity patterns.

On the other hand, large variations in the sparsity patterns realized post-projection can lead to instability in optimization, thus hampering the final performance of the model.
For instance, Top-KAST empirically benefits from the application of additional sparsity-inducing regularization on the parameters~[\citealp{jayakumar2020top} (Sec. 2.3), \citealp{schwarz2021powerpropagation}].
This issue motivates our use of a soft top-$k$ approximation as a mechanism for controlling the stability of our training iterates.

Each iteration of Spartan consists of the following two steps~(Algorithm~\ref{alg:spartan}): 
(1) an approximately $k$-sparse masking of the model parameters using the soft top-$k$ operator, and
(2) dual averaging-based parameter updates with the set of exactly $k$-sparse vectors as the feasible set.
This procedure aims to combine the advantages of Algorithms~\ref{alg:imp} and \ref{alg:da} within a single parameter update scheme.
Note that the Spartan parameter update is essentially identical to Algorithm~\ref{alg:da}, save for the inclusion of the intermediate soft top-$k$ masking step.

\begin{algorithm}[h]
    \caption{Spartan parameter update}
    \label{alg:spartan}
    \begin{algorithmic}[1]
        \Require parameters $\theta_t$, loss function $L(\theta)$, sparsity budget $k$, sharpness parameter $\beta$, step size $\eta_t$
        \Ensure parameters $\theta_{t+1}$
        \State $\sigma_k^\beta(\theta_t) = \theta_t \odot \mathrm{softtopk}(|\theta_t|, k, \beta)$ \Comment{apply soft masking}
        \State $\tilde{\theta}_t = \Pi_k\left(\sigma_k^\beta(\theta_t)\right)$ \Comment{project onto $k$-sparse set}
        \State $\theta_{t+1} = \theta_t - \eta_t \nabla \sigma_k^\beta(\theta_t) \nabla L(\tilde{\theta}_t)$ \Comment{compute dual averaging update}
        \State \textbf{return} $\theta_{t+1}$
    \end{algorithmic}
\end{algorithm}

In the following, we begin by describing the soft top-$k$ masking operation and address the issues of scalability and of applying soft top-$k$ masking to induce structured sparsity.
We then discuss how the update procedure outlined in Algorithm~\ref{alg:spartan} can be incorporated within a complete training loop.

\subsection{Soft Top-$k$ Masking}

Our soft top-$k$ masking scheme is based on the soft top-$k$ operator $\mathrm{softtopk}(v, k, \beta)$ described by \citet{xie2020differentiable}.
$\mathrm{softtopk}$ takes as input a vector $v\in \mathbb{R}^d$, a budget parameter $k > 0$ and a sharpness parameter $\beta \geq 0$, and outputs
$m \in [0, 1]^d$, a smoothed version of the top-$k$ indicator vector.
By using the magnitudes $|\theta|$ of the parameter vector as a measure of the ``value'' of each entry, we obtain a soft top-$k$ magnitude pruning operator $\sigma_k^\beta(\cdot)$
by masking the entries of the parameter vector with the output of the soft top-$k$ operator:
\begin{equation*}
    \sigma_k^\beta(\theta) \coloneqq \theta \odot \mathrm{softtopk}(|\theta|, k, \beta).
\end{equation*}

We introduce a mild generalization of the soft top-$k$ operator from \citet{xie2020differentiable} by incorporating a strictly positive \emph{cost vector} $c$.
In particular, we require that the output mask $m \in [0, 1]^d$ satisfies the budget constraint $c^T m = k$.
This abstraction is useful for modeling several notions of heterogeneous parameter costs within a 
model;
for instance, parameters that are repeatedly invoked within a network have a higher associated computational cost.
As a concrete example, \citet{evci2020rigging} observed that the FLOP count of a ResNet-50 model at a fixed global sparsity can vary by a factor of over $2\times$
due to differences in the individual sparsities of convolutional layers with differing output dimensions.

To derive this cost-sensitive soft top-$k$ operator, we begin by expressing the mask $m$ as the solution to the following linear program (LP):
\begin{align}
    \underset{m \in \mathbb{R}^d}{\rm maximize}\quad & v^T m \quad \text{subject to}\quad 0 \preceq m \preceq \one_d, \;\; c^T m = k.\label{eq:lp}
\end{align}

Deferring the derivation to the Appendix, we can rewrite this LP as the following optimal transportation (OT) problem:
\begin{align}
    \underset{Y \in \mathbb{R}_+^{d \times 2}}{\rm minimize}\quad & \sum_{ij} C_{ij} Y_{ij} \quad \text{subject to}\quad  Y \one_2 = c, \; \one_d Y = [k,\, \one^T_d c - k], \label{eq:ot}
\end{align}
with the cost matrix $C \coloneqq [-v/c, \, 0]$. We can recover the mask from $Y$ as $m_i = Y_{i1} / c_i$.
Now following \citep{xie2020differentiable,cuturi2013sinkhorn}, we introduce entropic regularization to obtain a smoothed version of the top-$k$ operator, with the degree of smoothing controlled by $\beta$:
\begin{align}
    \underset{Y \in \mathbb{R}_+^{d \times 2}}{\rm minimize}\quad & \sum_{ij} C_{ij} Y_{ij} - \frac{1}{\beta} H(Y) \label{eq:reg-ot} \\
    \text{subject to}\quad & Y \one_2 = c, \; \one_d Y = [k,\, \one^T_d c - k], \nonumber
\end{align}
where $H(Y) \coloneqq -\sum_{ij} Y_{ij} (\log Y_{ij} - 1)$ is the entropy of $Y$.
In this standard form, we can efficiently solve this regularized OT problem using the Sinkhorn-Knopp algorithm~\citep{cuturi2013sinkhorn,benamou2015iterative,cuturi2019differentiable}.

\begin{algorithm}[t]
    \caption{Soft top-$k$ forward pass}
    \label{alg:softtopk-forward}
    \begin{algorithmic}[1]
        \Require values~$v \in \mathbb{R}^d$, costs~$c \in \mathbb{R}_{++}^d$, budget~$k \in (0, \one_d^T c]$, sharpness parameter~$\beta \geq 0$, max.~iterations~$T$, tolerance~$\epsilon$, initial dual variable~$\mu_0$
        \Ensure mask~$m \in [0, 1]^d$, dual variables~$\mu \in \mathbb{R}$, $\nu \in \mathbb{R}^d$
        \State $z = \beta v / c, \;\; m_0 = \one_d$ \Comment{normalize values $v$ by costs $c$}
        \For{$t = 1, \dots, T$} 
            \State $\nu_t = \log c - \log(\one_d + \exp(z + \mu_{t-1}))$ \Comment{normalize mask entries}
            \State $\mu_t = \log k - \log\sum_{i}\exp(z_i + \nu_{t,i})$ \Comment{normalize sum to be equal to $k$}
            \State $m_t = \exp(z + \mu_t + \nu_t - \log c)$ \Comment{compute mask}
            \If{$|v^T(m_t - m_{t-1})| < \epsilon |v^T m_{t-1}|$} \Comment{check stopping criterion}
               \State \textbf{return} $m_t$, $\mu_t$, $\nu_t$
            \EndIf
        \EndFor
        \State \textbf{return}  $m_T$, $\mu_T$, $\nu_T$
    \end{algorithmic}
\end{algorithm}

\begin{algorithm}[t]
    \caption{Soft top-$k$ backward pass}
    \label{alg:softtopk-backward}
    \begin{algorithmic}[1]
        \Require gradient w.r.t. outputs $g \in \mathbb{R}^d$, mask $m \in[0,1]^d$, costs $c$, budget $k$, sharpness parameter $\beta$
        \Ensure gradient w.r.t. inputs
        \State $a_1 = \sum_{i} g_i m_i (1 - m_i), \;\; a_2 = \sum_{i} c_i m_i^2$
        \State \textbf{return}  $\beta m \odot (\one_d - m) \odot (g / c -  a_1 / (k - a_2))$ 
    \end{algorithmic}
\end{algorithm}

Algorithms~\ref{alg:softtopk-forward} and \ref{alg:softtopk-backward} describe the forward and backward passes of
the resulting cost-sensitive soft top-$k$ operator (see Appendices~\ref{sec:forward-derivation} and \ref{sec:backward-derivation} for their derivations).
The expression for the gradient in Algorithm~\ref{alg:softtopk-backward} follows from Theorem 3 in \citet{xie2020differentiable} with some algebraic simplification.
We remark that this closed-form backward pass is approximate in the sense that it assumes that we obtain the \emph{optimal}
dual variables $\mu, \nu$ in the forward pass; in practice, we do not encounter issues when using estimates of the
dual variables obtained with tolerance $\epsilon = 0.01$ and a maximum iteration count of $T=100$ in Algorithm~\ref{alg:softtopk-forward}.

\minihead{Scalability} Since we apply soft top-$k$ masking to high-dimensional parameters in each iteration of training,
it is important to minimize the computational overhead of Sinkhorn iteration in the forward pass.
We note that in order to compute the hard top-$k$ projection step in dual averaging (as in Algorithm~\ref{alg:spartan}), it is necessary to 
find the index $i_k$ of the $k$th largest entry of $|\theta|/c$.
Given the index $i_k$, we can use the heuristic initialization $\mu_0 = -\beta|\theta_{i_k}|/c_{i_k}$ for the dual variable $\mu$ to accelerate the convergence of Sinkhorn iteration. 
This heuristic is based on the observation that $|\theta_{i_k}|/c_{i_k}$ is the threshold value of the normalized value vector $|\theta|/c$ in \emph{hard} top-$k$ masking.
Concretely, we demonstrate in Sec.~\ref{sec:overhead} that our implementation of Spartan incurs a per-iteration runtime overhead of approximately 5\% over standard dense ResNet-50 training.

\minihead{Structured sparsity}
To implement block structured sparsity, we instantiate one mask variable per block and mask all parameters within that block
with the same mask variable.
To compute the mask, we use the sum of the magnitudes of the entries in each block as the corresponding value $v_i$.
In a standard pruning setting with uniform costs across all parameters, the corresponding cost $c_i$ is simply the total number 
of entries in the block.

\subsection{Training with Spartan}
\label{sec:training-spartan}

\minihead{Training schedule} When training with Spartan, we divide the training process into a warmup phase, an intermediate phase, and a fine-tuning phase.
In our experiments, the warmup phase consists of the first 20\% of epochs, the intermediate phase the next 60\%, and the fine-tuning phase the final 20\%.
In the warmup phase, we linearly anneal the global sparsity of the model from a value of $1$ at the start of training to the target level of sparsity.
Throughout the intermediate phase, we maintain the target sparsity level, and during the fine-tuning phase, we fix the sparsity mask to that used at the start of fine-tuning.
This training schedule is similar to those used in prior work~\citep{schwarz2021powerpropagation}.

\begin{figure}
    \centering
    \includegraphics[width=\textwidth]{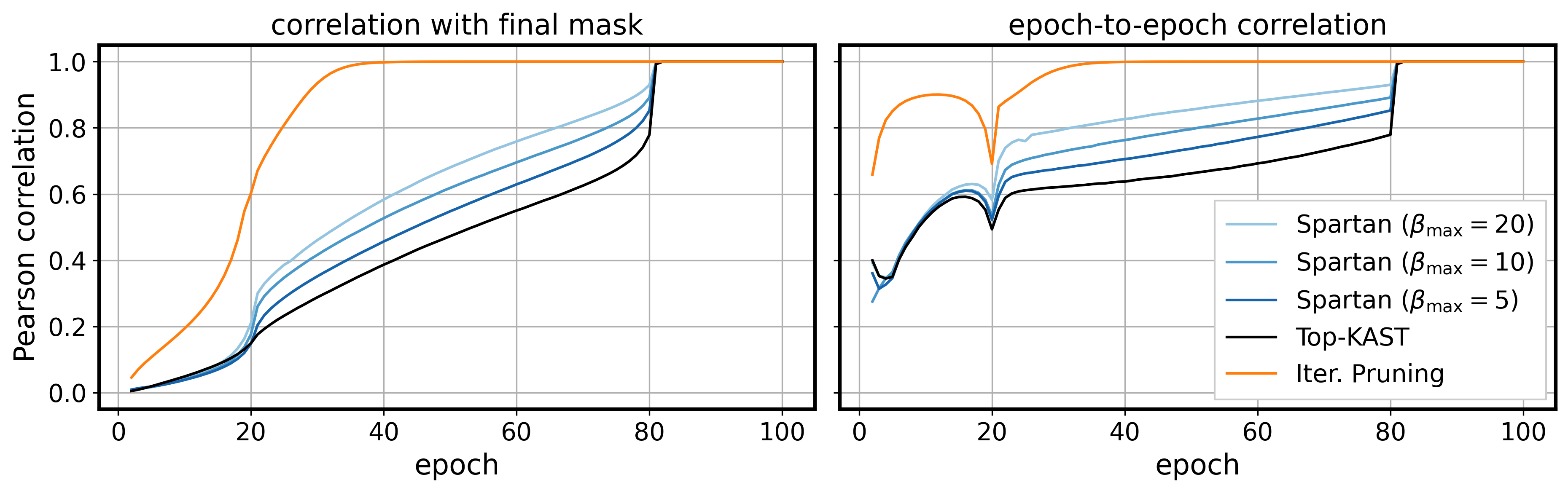}
    \vspace{-1.5em}
    \caption{\textbf{Spartan parameterizes an exploration-exploitation tradeoff.} 
    For each run of ResNet-50 training on ImageNet-1K, we plot Pearson correlation coefficients between the sparsity mask at the end of each epoch and 
    the mask obtained at the end of training (left), and between sparsity masks at the end of subsequent epochs (right). 
    Kinks in the correlation curves at epochs 20 and 80 are respectively due to the end of sparsity annealing and the start of fine-tuning with fixed masks
    (see Section~\ref{sec:training-spartan} for details on the training schedule).
    }
    \label{fig:correlation}
\end{figure}

\minihead{Exploration vs. exploitation as a function of $\beta$}
From the start of training up to the start of the fine-tuning phase, we linearly ramp the sharpness parameter $\beta$ from an initial value
of $1$ to the final value $\beta_\mathrm{max}$.
We interpret the spectrum of updates parameterized by $\beta$ as being characteristic of an exploration-exploitation tradeoff.
In Figure~\ref{fig:correlation}, we illustrate this phenomenon by plotting Pearson correlation coefficients between sparsity masks at
different stages of training.
We observe that iterative magnitude pruning converges on the final mask relatively early in training, which is indicative of
insufficient exploration of different sparsity patterns---consequently, this model achieves lower validation accuracy than the remaining
models.
The three models trained with Spartan each ramp $\beta$ from $1$ at the start of training to $\beta_\mathrm{max}$ at epoch 80. 
The value of $\beta_\mathrm{max}$ correlates well both with the Pearson correlation with the final mask, and with the Pearson correlation 
between masks at the end of subsequent epochs.
Spartan thus interpolates between the high-exploration regime of standard dual averaging and the low-exploration regime of iterative magnitude pruning.
In this example, the intermediate setting $\beta_\mathrm{max} = 10$ achieves the highest top-1 validation accuracy of $74.6\%$, with
Top-KAST at $73.5\%$ and IMP at $68.2\%$.

\section{Empirical Evaluation}
\label{sec:experiments}

In this section, we report the results of our sparse training experiments on the ImageNet-1K dataset with 
two standard architectures: ResNet-50~\citep{he2016deep} and ViT-B/16~\citep{dosovitskiy2020image}, consisting of 25.6M and 86.4M parameters respectively.
On ResNet-50, we evaluate only unstructured pruning, whereas on ViT-B/16, we evaluate both unstructured and block structured
pruning.
We subsequently present empirical studies of the sensitivity of Spartan to the value of $\beta$, the effect of running Spartan without the dual averaging step,
and the computational overhead of soft top-$k$ masking.

\begin{table}
    \caption{Top-1 accuracies on ImageNet-1K validation set with fully dense training.}
    \label{table:r50-baselines}
    \centering
    \resizebox{0.8\textwidth}{!}{
    \begin{tabular}{rcccrcc}
        \cmidrule[\heavyrulewidth]{1-3} \cmidrule[\heavyrulewidth]{5-7}
        \textbf{Method} & \textbf{Epochs} & \textbf{Accuracy (\%)} && \textbf{Method} & \textbf{Epochs} & \textbf{Accuracy (\%)}\\
        \cmidrule{1-3} \cmidrule{5-7}
        \multirow{3}{*}{ResNet-50}
        & $100$ & \res{77.08}{0.11} && \multirow{3}{*}{ViT-B/16} & \multirow{3}{*}{\begin{tabular}{c}$100$\\[\ExtraSep] $300$\end{tabular}} & \multirow{3}{*}{\begin{tabular}{c}\res{77.78}{0.22} \\[\ExtraSep] \bres{80.06}{0.11}\end{tabular}} \\
        & $200$ & \bres{77.46}{0.09} &&                          &                   &  \\
        & $400$ & \res{77.17}{0.04} && \\
        \cmidrule[\heavyrulewidth]{1-3} \cmidrule[\heavyrulewidth]{5-7}
    \end{tabular}
    }
\end{table}

\subsection{ImageNet-1K Classification}

\begin{table}
    \caption{ResNet-50 top-1 accuracies on ImageNet-1K validation set at varying levels of sparsity and epochs of training.
    When available, we report means and standard deviations over 3 trials.
    }
    \label{table:r50-imagenet}
    \centering
    \resizebox{\linewidth}{!}{
    \begin{tabular}{crcccccc}
        \cmidrule[\heavyrulewidth]{2-8}
        &  &  & \multicolumn{5}{c}{\textbf{Sparsity}} \\
        \cmidrule(lr){4-8}
        & \textbf{Method} & \textbf{Epochs} & \multicolumn{1}{c}{\textbf{80\%}} & \multicolumn{1}{c}{\textbf{90\%}} & \multicolumn{1}{c}{\textbf{95\%}} & \multicolumn{1}{c}{\textbf{97.5\%}} & \multicolumn{1}{c}{\textbf{99\%}} \\
        \cmidrule{2-8}
        \multirow{2}{*}{\citep{evci2020rigging}} 
        & \multirow{2}{*}{RigL {\small (ERK)}} & $100$ & \res{75.1}{0.05} & \res{73.0}{0.04} & \res{69.7}{0.17} & - & - \\
        &             & $500$ & \res{77.1}{0.06} & \res{76.4}{0.05} & \res{74.5}{0.09} & - & - \\
        \cmidrule{2-8}
        \citep{kusupati2020soft}
        & STR$^\dagger$ & $100$ & $76.19$ & $74.31$ & $70.40$ & $62.84$ & $51.82$ \\
        \cmidrule{2-8}
        \citep{zhou2021effective}
        & ProbMask & $100$ & - & $74.68$ & $71.50$ & $66.83^{\ddagger}$ & $61.07$ \\
        \cmidrule{2-8}
        \citep{zhang2022optimizing}
        & OptG & $100$ & - & $74.28$ & $72.38$ & - & $62.10$ \\
        \cmidrule{2-8}
        \multirow{7}{*}{\citep{schwarz2021powerpropagation}}
        & IMP & $100$ & \res{75.3}{0.07} & \res{73.7}{0.14} & \res{70.6}{0.05} & - & - \\
        \cmidrule(lr){4-8}
        & Top-KAST & $100$ & \res{76.08}{0.02} & \res{75.13}{0.03} & \res{73.19}{0.02} & - & - \\
        & {\small (with PP)} & $100$ & \res{76.24}{0.07} & \res{75.23}{0.02} & \res{73.25}{0.02} & - & - \\
        & {\small (with ERK)} & $100$ & \res{76.42}{0.03} & \res{75.51}{0.05} & - & - & - \\
        \cmidrule(lr){4-8}
        & {\small (with PP} & $100$ & \res{76.76}{0.08} & \res{75.74}{0.08} & - & - & - \\
        & {\small \& ERK)}  & $200$ & \res{77.51}{0.03} & \res{76.94}{0.10} & - & - & - \\
        &                  & $300$ & \res{77.64}{0.05} & \res{77.16}{0.19} & - & - & - \\
        \cmidrule{2-8}
        & \multirow{3}{*}{Top-KAST} & $100$ & \res{76.66}{0.04} & \res{75.48}{0.15} & \res{73.51}{0.16} & \res{70.23}{0.05} & - \\
        &         & $200$ & \res{77.55}{0.08} & \res{76.84}{0.11} & \res{75.20}{0.11} & \res{71.72}{0.06} & - \\
        &         & $400$ & \bres{77.75}{0.09} & \res{77.37}{0.07} & \res{75.90}{0.04} & \res{72.58}{0.13} & - \\
        \cmidrule(lr){4-8}
        & \multirow{3}{*}{Spartan}  & $100$ & \bres{76.89}{0.09} & \bres{76.17}{0.10} & \bres{74.68}{0.24} & \bres{71.95}{0.11} & \bres{63.87}{0.12} \\
        &         & $200$ & \bres{77.56}{0.19} & \bres{77.06}{0.13} & \bres{75.92}{0.01} & \bres{73.41}{0.15} & \res{65.77}{0.03} \\
        &         & $400$ & \res{77.57}{0.08} & \bres{77.40}{0.06} & \bres{76.48}{0.20} & \bres{74.15}{0.10} & \res{66.80}{0.03} \\
        \cmidrule[\heavyrulewidth]{2-8}
        & \multicolumn{7}{l}{\footnotesize $^\dagger$Reported accuracies for models closest to the listed sparsity level. $^\ddagger$Model trained at 98\% sparsity.}
    \end{tabular}
    }  
\end{table}

In all our experiments, we run Spartan with the training schedule described in Section~\ref{sec:training-spartan}.
We train and evaluate our models on the ImageNet-1K dataset with the standard training-validation split and report means and standard deviations over 3 independent trials.
We provide additional details on our experimental setup in the Appendix.

\minihead{ResNet-50 experimental setup} 
For consistency with our baselines, we use standard data augmentation with horizontal flips and random crops at $224\times 224$ resolution.
For all Spartan runs, we use $\beta_\mathrm{max} = 10$, which we selected based on models trained at 95\% accuracy.
We sparsify the parameters of all linear and convolutional layers with a global sparsity budget, excluding bias parameters and the parameters of batch normalization layers.
Our baselines are iterative magnitude pruning~\citep{zhu2018prune}, RigL with the Erdos-Renyi-Kernel (ERK) sparsity distribution~\citep{evci2020rigging},
Soft Threshold Weight Reparameterization (STR)~\citep{kusupati2020soft}, probabilistic masking (ProbMask)~\citep{zhou2021effective},
OptG~\citep{zhang2022optimizing}, and Top-KAST with Powerpropagation and ERK~\citep{jayakumar2020top,schwarz2021powerpropagation}.
We additionally re-run the most performant baseline method, Top-KAST, using a reimplementation in our own codebase.
For Top-KAST, we exclude the first convolutional layer from pruning (following \citep{schwarz2021powerpropagation,evci2020rigging})
and we use fully dense backward passes (\emph{i.e.}, with a backward sparsity of 0), since this setting achieved the highest accuracy in prior work~\citep{jayakumar2020top}.
We use mixed precision training with a batch size of 4096 on 8 NVIDIA A100 GPUs.

\begin{table}[t]
    \caption{ViT-B/16 top-1 accuracies on ImageNet-1K validation set at 90\% sparsity
    with unstructured sparsity and block structured pruning of attention layers.}
    \label{table:vit-imagenet}
    \centering
    \resizebox{0.75\linewidth}{!}{
        \begin{tabular}{rcccc}
            \toprule
             & & \multicolumn{3}{c}{\textbf{Sparsity Structure}} \\
             \cmidrule{3-5}
             \textbf{Method} & \textbf{Epochs} & \textbf{Unstructured} & $\mathbf{16\times16}$ \textbf{blocks} & $\mathbf{32\times32}$ \textbf{blocks} \\
            \midrule
            \multirow{2}{*}{Top-KAST} & $100$ & \res{78.05}{0.07} & \res{67.69}{0.23} & \res{64.79}{0.14} \\
             & $300$ &  \res{80.86}{0.03} & \res{74.11}{0.20} & \res{70.67}{0.96} \\
            \cmidrule(lr){3-5}
            \multirow{2}{*}{Spartan} & $100$ & \bres{79.88}{0.16} & \bres{75.62}{0.07} & \bres{74.50}{0.23} \\
            & $300$ & \bres{81.18}{0.04} & \bres{79.12}{0.16} & \bres{78.43}{0.10} \\
            \bottomrule
        \end{tabular}
    } 
\end{table}

\minihead{ViT experimental setup} 
We use the ViT-B architecture with $16\times 16$ patches at $224\times 224$ input resolution.
We augment the training data using RandAugment~\citep{cubuk2020randaugment}, MixUp~\citep{zhang2018mixup} and CutMix~\citep{yun2019cutmix}.
Our ViT models are trained from random initialization, without any pretraining.
We set $\beta_\mathrm{max} = 20$ for Spartan with unstructured sparsity, and 
$\beta_\mathrm{max} = 320$ and $\beta_\mathrm{max} = 640$ for Spartan with $16\times 16$ and $32\times 32$ blocks respectively. 
These latter settings are the values of $\beta_\mathrm{max}$ for the 
unstructured case scaled up by factors of $16$ and $32$---since each block averages the magnitudes of $B^2$ entries,
we expect the variance of the block values to correspondingly decrease by a factor of $B^2$, and we thus compensate by scaling up $\beta$ by $B$.
In the block structured case, we exclude the input convolutional layer and the output classification head from pruning since
their parameter dimensions are not divisible by the block size.
We use mixed precision training with a batch size of 4096 on 16 NVIDIA A100 GPUs across 2 nodes.

\minihead{Results} 
Table~\ref{table:r50-baselines} lists the top-1 validation accuracies achieved by fully dense ResNet-50 and ViT-B/16 models.
Table~\ref{table:r50-imagenet} reports validation accuracies for ResNet-50 models at 80\%, 90\%, 95\%, 97.5\% and 99\% sparsity,
and Table~\ref{table:vit-imagenet} reports validation accuracies for ViT at 90\% sparsity.
In the Appendix, we report additional measurements of inference FLOP costs for our sparse models and the results of experiments with FLOP-sensitive pruning.

For ResNet-50 models, we find that Spartan outperforms all our baselines across all training durations at sparsity levels of 90\% and above.
In particular, Spartan achieves a mean top-1 accuracy within $1\%$ of fully dense training at 95\% sparsity.
We observe that additional epochs of sparse training consistently improves the final generalization
accuracy; in contrast, validation accuracy peaks at 200 epochs for dense training.
This trend persists at 800 training epochs, where Spartan  achieves $67.18\pm 0.13$ top-1 accuracy at 99\% sparsity.
For the Top-KAST baseline, we omit results at 99\% sparsity due to training instability.
We note that the accuracy improvements in the Spartan-trained ResNet-50 models relative to Top-KAST are not a result of increased FLOP counts at a given level of sparsity,
as can be seen from the FLOP measurements reported in Appendix~\ref{sec:flop-measurements}.

For ViT-B/16 models, Spartan outperforms Top-KAST for both unstructured and block structured pruning.
We observe a particularly large improvement over Top-KAST in the block structured case, where Spartan improves absolute validation accuracy 
by $7.8\%$ for $32\times 32$ block structured pruning.
For unstructured pruning, Spartan achieves comparable accuracy to SViTE~\citep{chen2021chasing} ($81.2\%$ for Spartan vs. $81.3\%$ for SViTE),
but with $30\%$ higher sparsity ($90\%$ for Spartan vs. $60\%$ for SViTE).
In exchange for a $0.6\%$ reduction in accuracy relative to DeiT-B~\citep{touvron2021training}, Spartan reduces model storage cost by $10\times$ and the FLOP cost of inference by $7.4\times$.

\subsection{Sensitivity and Ablation Analysis}

\begin{figure}
    \centering
    \vspace{-0.2em}
    \includegraphics[width=0.9\textwidth]{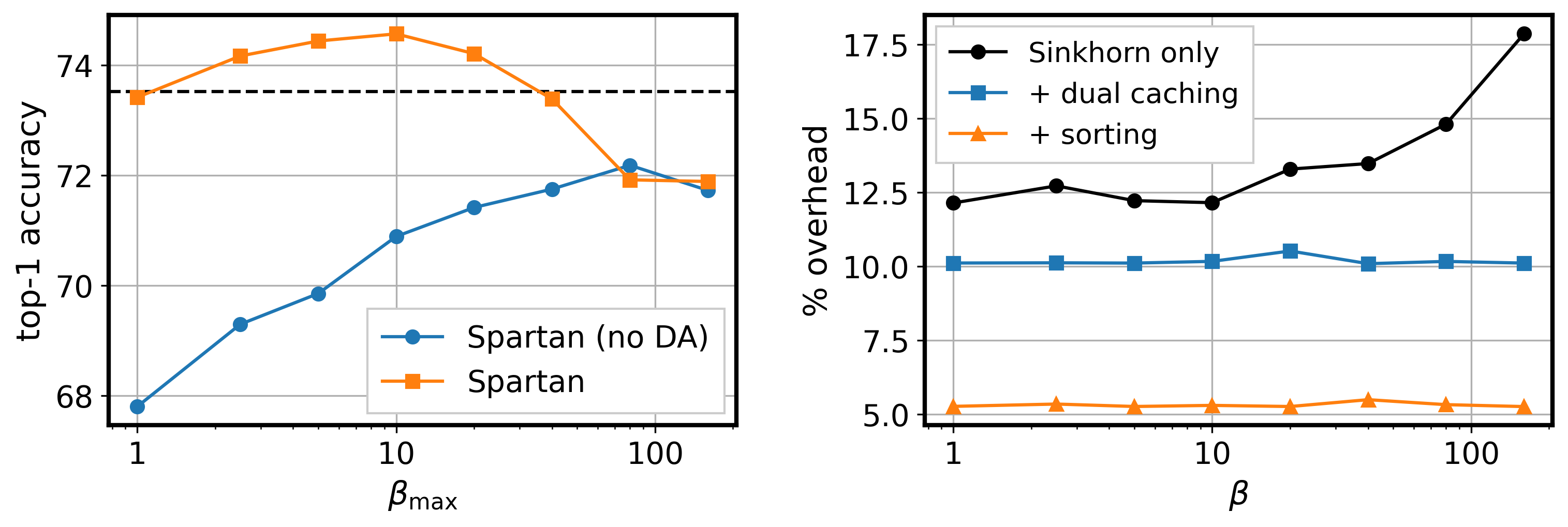}
    \vspace{-0.6em}
    \caption{\textbf{Effect of $\beta$ and dual averaging (left).} Top-1 ImageNet-1K validation accuracies for Spartan with and without dual averaging (DA) and with varying $\beta_\mathrm{max}$ at 95\% sparsity.
    \textbf{Computational overhead (right).} Percentage increase in per-iteration wall clock runtime over dense training for Spartan with standard Sinkhorn iteration,
    Sinkhorn with dual caching, and Sinkhorn with sorting.
    }
    \vspace{-0.5em}
    \label{fig:betas}
\end{figure}

Figure~\ref{fig:betas}~(left) shows the effect of ablating the dual averaging step in Spartan---\emph{i.e.}, omitting hard top-$k$ projection in the forward pass---over
a range of settings of $\beta_\mathrm{max}$ for 95\% sparse ResNet-50 models trained for 100 epochs.
The dashed line shows top-1 accuracy with Top-KAST.
For Spartan training without dual averaging, we compute a hard top-$k$ mask at the end of epoch 80, and as with 
standard Spartan training, we fix this mask until the end of training.
In the absence of top-$k$ projection, we find that accuracy increases with increasing $\beta_\mathrm{max}$ up to $\beta_\mathrm{max} = 80$;
at lower settings of $\beta_\mathrm{max}$, the higher approximation error of soft masking is detrimental to the final accuracy of the model. 
In contrast, the use of top-$k$ projection in the forward pass mitigates this mismatch between training and inference and
improves the final accuracy of the sparse model.

\subsection{Computational Overhead}
\label{sec:overhead}

We evaluate the computational overhead of Spartan over standard dense training by 
measuring wall clock training times for a ResNet-50 model (Figure~\ref{fig:betas}, right).
Our benchmark measures runtime on a single NVIDIA A100 GPU over 50 iterations of training with a batch size of 256.
We use random inputs and labels to avoid incurring data movement costs.
We compare three approaches: standard Sinkhorn iteration, Sinkhorn iteration with
the dual variable $\mu$ initialized using its final value from the previous iteration (``dual caching''), 
and Sinkhorn iteration with $\mu$ initialized using the value $-\beta|\theta_{i_k}|/c_{i_k}$, where
we compute the index $i_k$ of the $k$th largest entry of the normalized value vector $|\theta|/c$ by sorting its entries in each iteration (``sorting'').

Standard Sinkhorn iteration incurs higher overheads as $\beta$ increases---this is due to the
additional iterations required to reach convergence as the regularized OT problem more closely approximates the original LP.
We find that dual caching and sorting both prevent this growth in runtime over the range 
of $\beta$ values that we tested. In our remaining experiments, we use the simple sorting-based approach since it 
corresponds to the lowest relative overhead over standard dense training (approximately 5\%).
We note that this relative overhead decreases as the batch size increases since the cost of computing the mask is independent of the batch size.
Spartan is therefore best suited to training with large batch sizes since we amortize the cost of mask updates over the size of the batch.

\section{Discussion}
\label{sec:discussion}

\minihead{Extensions}
As an alternative to magnitude-based pruning, we may also apply Spartan in conjunction with \emph{learned value parameters}, as in methods such as Movement Pruning~\citep{sanh2020movement}.
In this approach, we would compute masks using a set of auxiliary parameters $\phi \in \mathbb{R}^d$ instead of the magnitudes $|\theta|$:
$\sigma_k^\beta(\theta; \phi) = \theta \odot \mathrm{softtopk}(\phi, k, \beta)$, and similarly for hard projection.
We remark that while this variant requires additional memory during training to store the value parameters, there is 
no additional cost during inference since the sparsity mask is fixed at the end of training.   

\minihead{Limitations}
Since Spartan retains a dense parameter vector and computes dense backward passes during training, it incurs 
higher memory and computational costs in each iteration than methods like RigL~\citep{evci2020rigging} that use both a sparse parameters and sparse backward passes.
Nevertheless, we note that in terms of \emph{total} computational or energy cost over a full training run, Spartan may
remain a competitive option as it requires fewer iterations of training to reach a given accuracy threshold relative to
 these sparse-to-sparse methods. However, we do not compare total training FLOP costs in our empirical evaluation.

A further limitation is that cost-sensitive pruning with Spartan is only compatible with relatively crude linear cost models of the form $c^T m$, where $c$ is the cost vector and $m$ is the mask.
This restriction arises due to the requirements of the regularized OT formulation used to compute the soft top-$k$ mask.
In particular, this limitation precludes the use of cost models involving interaction terms such as those arising from spatially coherent sparsity patterns.
For example, the cost models that we consider here cannot encode the notion that coherently pruning an entire row or column of a parameter matrix will yield more cost savings than pruning random entries of the matrix.

\minihead{Optimization with Dual Averaging}
As discussed in Section~\ref{sec:method}, Spartan and Top-KAST are both applications of the dual averaging method~\citep{nesterov2009primal,xiao2009dual}.
The empirically observed effectiveness of dual averaging in sparse training is reminiscent of its success in a related domain
where continuous optimization is complicated by discrete constraints: neural network quantization.
In this area, the BinaryConnect algorithm~\citep{courbariaux2015binaryconnect}, which trains binary neural networks with parameters in $\{\pm 1\}^d$, 
is considered to be a foundational method. 
As observed by \citet{bai2018proxquant}, BinaryConnect is itself a special case of dual averaging, and the theoretical underpinnings of this connection to dual averaging were further analyzed by \citet{dockhorn2021demystifying}.
These parallels with neural network quantization suggest that similar ideas may well apply towards deepening our understanding of sparse learning.

\minihead{Societal Impacts}
Inference with deep neural network models is a computationally intensive process. 
At present, the total energy footprint associated with serving these models in production 
is expected to continue growing in tandem with the rising prevalence of large transformer-based architectures
in vision and NLP applications.
Research towards improving the energy efficiency of deep neural networks is therefore an 
important counterbalancing force against increasing resource usage by these models.
The development of sparse learning algorithms is particularly relevant to these efforts, and 
we expect that the impact of these approaches will further increase as sparsity-aware hardware acceleration becomes 
more widely available.

\section{Conclusions \& Future Work}

In this work, we describe a sparse learning algorithm that interpolates between two parameter update schemes:
standard stochastic gradient updates with hard masking and the dual averaging method.
We show that there exists an intermediate regime between these two methods that yields improved generalization accuracy for
sparse convolutional and transformer models, particularly at higher levels of sparsity.
While we have demonstrated promising empirical results with our proposed method, the learning dynamics of stochastic optimization
for deep networks under sparsity constraints remains relatively poorly understood from a theoretical standpoint.
There thus exists ample opportunity for further work towards better understanding sparse learning algorithms, which may in turn inspire
future algorithmic advances in this area.

\section*{Acknowledgements and Disclosure of Funding}

We are grateful to Trevor Gale for his feedback on this work.
We also thank our anonymous reviewers for their valuable suggestions on improving the manuscript.
This work was funded by Meta.

\bibliography{references}

\begin{thebibliography}{48}
\providecommand{\natexlab}[1]{#1}
\providecommand{\url}[1]{\texttt{#1}}
\expandafter\ifx\csname urlstyle\endcsname\relax
  \providecommand{\doi}[1]{doi: #1}\else
  \providecommand{\doi}{doi: \begingroup \urlstyle{rm}\Url}\fi

\bibitem[Azarian et~al.(2020)Azarian, Bhalgat, Lee, and
  Blankevoort]{azarian2020learned}
Kambiz Azarian, Yash Bhalgat, Jinwon Lee, and Tijmen Blankevoort.
\newblock Learned threshold pruning.
\newblock \emph{arXiv preprint arXiv:2003.00075}, 2020.

\bibitem[Bai et~al.(2018)Bai, Wang, and Liberty]{bai2018proxquant}
Yu~Bai, Yu-Xiang Wang, and Edo Liberty.
\newblock {ProxQuant: Quantized Neural Networks via Proximal Operators}.
\newblock In \emph{International Conference on Learning Representations}, 2018.

\bibitem[Bellec et~al.(2018)Bellec, Kappel, Maass, and
  Legenstein]{bellec2018deep}
Guillaume Bellec, David Kappel, Wolfgang Maass, and Robert Legenstein.
\newblock {Deep Rewiring: Training very sparse deep networks}.
\newblock In \emph{International Conference on Learning Representations}, 2018.

\bibitem[Benamou et~al.(2015)Benamou, Carlier, Cuturi, Nenna, and
  Peyr{\'e}]{benamou2015iterative}
Jean-David Benamou, Guillaume Carlier, Marco Cuturi, Luca Nenna, and Gabriel
  Peyr{\'e}.
\newblock {Iterative Bregman projections for regularized transportation
  problems}.
\newblock \emph{SIAM Journal on Scientific Computing}, 37\penalty0 (2), 2015.

\bibitem[Bengio et~al.(2013)Bengio, L{\'e}onard, and
  Courville]{bengio2013estimating}
Yoshua Bengio, Nicholas L{\'e}onard, and Aaron Courville.
\newblock Estimating or propagating gradients through stochastic neurons for
  conditional computation.
\newblock \emph{arXiv preprint arXiv:1308.3432}, 2013.

\bibitem[Chavan et~al.(2022)Chavan, Shen, Liu, Liu, Cheng, and
  Xing]{chavan2022vision}
Arnav Chavan, Zhiqiang Shen, Zhuang Liu, Zechun Liu, Kwang-Ting Cheng, and Eric
  Xing.
\newblock Vision transformer slimming: Multi-dimension searching in continuous
  optimization space.
\newblock \emph{arXiv preprint arXiv:2201.00814}, 2022.

\bibitem[Chen et~al.(2021)Chen, Cheng, Gan, Yuan, Zhang, and
  Wang]{chen2021chasing}
Tianlong Chen, Yu~Cheng, Zhe Gan, Lu~Yuan, Lei Zhang, and Zhangyang Wang.
\newblock Chasing sparsity in vision transformers: An end-to-end exploration.
\newblock In \emph{Advances in Neural Information Processing Systems}, 2021.

\bibitem[Collins and Kohli(2014)]{collins2014memory}
Maxwell~D Collins and Pushmeet Kohli.
\newblock Memory bounded deep convolutional networks.
\newblock \emph{arXiv preprint arXiv:1412.1442}, 2014.

\bibitem[Courbariaux et~al.(2015)Courbariaux, Bengio, and
  David]{courbariaux2015binaryconnect}
Matthieu Courbariaux, Yoshua Bengio, and Jean-Pierre David.
\newblock {BinaryConnect: Training deep neural networks with binary weights
  during propagations}.
\newblock In \emph{Advances in Neural Information Processing Systems}, 2015.

\bibitem[Cubuk et~al.(2020)Cubuk, Zoph, Shlens, and Le]{cubuk2020randaugment}
Ekin~D Cubuk, Barret Zoph, Jonathon Shlens, and Quoc~V Le.
\newblock Randaugment: Practical automated data augmentation with a reduced
  search space.
\newblock In \emph{IEEE/CVF Conference on Computer Vision and Pattern
  Recognition Workshops}, 2020.

\bibitem[Cuturi(2013)]{cuturi2013sinkhorn}
Marco Cuturi.
\newblock Sinkhorn distances: Lightspeed computation of optimal transport.
\newblock In \emph{Advances in Neural Information Processing Systems}, 2013.

\bibitem[Cuturi et~al.(2019)Cuturi, Teboul, and Vert]{cuturi2019differentiable}
Marco Cuturi, Olivier Teboul, and Jean-Philippe Vert.
\newblock Differentiable ranking and sorting using optimal transport.
\newblock In \emph{Advances in Neural Information Processing Systems}, 2019.

\bibitem[Dettmers and Zettlemoyer(2019)]{dettmers2019sparse}
Tim Dettmers and Luke Zettlemoyer.
\newblock Sparse networks from scratch: Faster training without losing
  performance.
\newblock \emph{arXiv preprint arXiv:1907.04840}, 2019.

\bibitem[Dockhorn et~al.(2021)Dockhorn, Yu, Sari, Zolnouri, and
  Partovi~Nia]{dockhorn2021demystifying}
Tim Dockhorn, Yaoliang Yu, Eyy{\"u}b Sari, Mahdi Zolnouri, and Vahid
  Partovi~Nia.
\newblock Demystifying and generalizing binaryconnect.
\newblock 2021.

\bibitem[Dosovitskiy et~al.(2021)Dosovitskiy, Beyer, Kolesnikov, Weissenborn,
  Zhai, Unterthiner, Dehghani, Minderer, Heigold, Gelly,
  et~al.]{dosovitskiy2020image}
Alexey Dosovitskiy, Lucas Beyer, Alexander Kolesnikov, Dirk Weissenborn,
  Xiaohua Zhai, Thomas Unterthiner, Mostafa Dehghani, Matthias Minderer, Georg
  Heigold, Sylvain Gelly, et~al.
\newblock An image is worth 16x16 words: Transformers for image recognition at
  scale.
\newblock In \emph{International Conference on Learning Representations}, 2021.

\bibitem[Evci et~al.(2020)Evci, Gale, Menick, Castro, and
  Elsen]{evci2020rigging}
Utku Evci, Trevor Gale, Jacob Menick, Pablo~Samuel Castro, and Erich Elsen.
\newblock Rigging the lottery: Making all tickets winners.
\newblock In \emph{International Conference on Machine Learning}, 2020.

\bibitem[Gale et~al.(2019)Gale, Elsen, and Hooker]{gale2019state}
Trevor Gale, Erich Elsen, and Sara Hooker.
\newblock The state of sparsity in deep neural networks.
\newblock \emph{arXiv e-prints}, arXiv:1902.09574, 2019.

\bibitem[Gray et~al.(2017)Gray, Radford, and Kingma]{gray2017gpu}
Scott Gray, Alec Radford, and Diederik~P Kingma.
\newblock {GPU kernels for block-sparse weights}.
\newblock \emph{arXiv preprint arXiv:1711.09224}, 2017.

\bibitem[Han et~al.(2015)Han, Pool, Tran, and Dally]{han2015learning}
Song Han, Jeff Pool, John Tran, and William Dally.
\newblock Learning both weights and connections for efficient neural network.
\newblock \emph{Advances in Neural Information Processing Systems}, 2015.

\bibitem[Hassibi and Stork(1992)]{hassibi1992second}
Babak Hassibi and David Stork.
\newblock Second order derivatives for network pruning: Optimal brain surgeon.
\newblock \emph{Advances in Neural Information Processing Systems}, 1992.

\bibitem[He et~al.(2016)He, Zhang, Ren, and Sun]{he2016deep}
Kaiming He, Xiangyu Zhang, Shaoqing Ren, and Jian Sun.
\newblock {Deep Residual Learning for Image Recognition}.
\newblock In \emph{IEEE Conference on Computer Vision and Pattern Recognition},
  2016.

\bibitem[Hoefler et~al.(2021)Hoefler, Alistarh, Ben-Nun, Dryden, and
  Peste]{hoefler2021sparsity}
Torsten Hoefler, Dan Alistarh, Tal Ben-Nun, Nikoli Dryden, and Alexandra Peste.
\newblock Sparsity in deep learning: Pruning and growth for efficient inference
  and training in neural networks.
\newblock \emph{Journal of Machine Learning Research}, 2021.

\bibitem[Janowsky(1989)]{janowsky1989pruning}
Steven~A Janowsky.
\newblock Pruning versus clipping in neural networks.
\newblock \emph{Physical Review A}, 1989.

\bibitem[Jayakumar et~al.(2020)Jayakumar, Pascanu, Rae, Osindero, and
  Elsen]{jayakumar2020top}
Siddhant Jayakumar, Razvan Pascanu, Jack Rae, Simon Osindero, and Erich Elsen.
\newblock {Top-KAST: Top-K Always Sparse Training}.
\newblock \emph{Advances in Neural Information Processing Systems}, 2020.

\bibitem[Kusupati et~al.(2020)Kusupati, Ramanujan, Somani, Wortsman, Jain,
  Kakade, and Farhadi]{kusupati2020soft}
Aditya Kusupati, Vivek Ramanujan, Raghav Somani, Mitchell Wortsman, Prateek
  Jain, Sham Kakade, and Ali Farhadi.
\newblock Soft threshold weight reparameterization for learnable sparsity.
\newblock In \emph{International Conference on Machine Learning}, 2020.

\bibitem[Lagunas et~al.(2021)Lagunas, Charlaix, Sanh, and
  Rush]{lagunas2021block}
Fran{\c{c}}ois Lagunas, Ella Charlaix, Victor Sanh, and Alexander~M Rush.
\newblock Block pruning for faster transformers.
\newblock In \emph{Proceedings of the 2021 Conference on Empirical Methods in
  Natural Language Processing}, 2021.

\bibitem[LeCun et~al.(1989)LeCun, Denker, and Solla]{lecun1989optimal}
Yann LeCun, John Denker, and Sara Solla.
\newblock Optimal brain damage.
\newblock \emph{Advances in Neural Information Processing Systems}, 2, 1989.

\bibitem[Lee et~al.(2018)Lee, Ajanthan, and Torr]{lee2018snip}
Namhoon Lee, Thalaiyasingam Ajanthan, and Philip Torr.
\newblock {SNIP: Single-shot Network Pruning based on Connection Sensitivity}.
\newblock In \emph{International Conference on Learning Representations}, 2018.

\bibitem[Louizos et~al.(2018)Louizos, Welling, and Kingma]{louizos2018learning}
C~Louizos, M~Welling, and DP~Kingma.
\newblock {Learning sparse neural networks through $L_0$ regularization}.
\newblock In \emph{International Conference on Learning Representations}, 2018.

\bibitem[Luo and Wu(2020)]{luo2020autopruner}
Jian-Hao Luo and Jianxin Wu.
\newblock Autopruner: An end-to-end trainable filter pruning method for
  efficient deep model inference.
\newblock \emph{Pattern Recognition}, 2020.

\bibitem[Mocanu et~al.(2018)Mocanu, Mocanu, Stone, Nguyen, Gibescu, and
  Liotta]{mocanu2018scalable}
Decebal~Constantin Mocanu, Elena Mocanu, Peter Stone, Phuong~H Nguyen,
  Madeleine Gibescu, and Antonio Liotta.
\newblock Scalable training of artificial neural networks with adaptive sparse
  connectivity inspired by network science.
\newblock \emph{Nature Communications}, 2018.

\bibitem[Mostafa and Wang(2019)]{mostafa2019parameter}
Hesham Mostafa and Xin Wang.
\newblock Parameter efficient training of deep convolutional neural networks by
  dynamic sparse reparameterization.
\newblock In \emph{International Conference on Machine Learning}, 2019.

\bibitem[Mozer and Smolensky(1988)]{mozer1988skeletonization}
Michael~C Mozer and Paul Smolensky.
\newblock Skeletonization: A technique for trimming the fat from a network via
  relevance assessment.
\newblock In \emph{Advances in Neural Information Processing Systems}, 1988.

\bibitem[Nesterov(2009)]{nesterov2009primal}
Yurii Nesterov.
\newblock Primal-dual subgradient methods for convex problems.
\newblock \emph{Mathematical Programming}, 2009.

\bibitem[Sanh et~al.(2020)Sanh, Wolf, and Rush]{sanh2020movement}
Victor Sanh, Thomas Wolf, and Alexander Rush.
\newblock Movement pruning: Adaptive sparsity by fine-tuning.
\newblock In \emph{Advances in Neural Information Processing Systems}, 2020.

\bibitem[Savarese et~al.(2020)Savarese, Silva, and Maire]{savarese2020winning}
Pedro Savarese, Hugo Silva, and Michael Maire.
\newblock Winning the lottery with continuous sparsification.
\newblock \emph{Advances in Neural Information Processing Systems}, 2020.

\bibitem[Schwarz et~al.(2021)Schwarz, Jayakumar, Pascanu, Latham, and
  Teh]{schwarz2021powerpropagation}
Jonathan Schwarz, Siddhant Jayakumar, Razvan Pascanu, Peter Latham, and Yee
  Teh.
\newblock Powerpropagation: A sparsity inducing weight reparameterisation.
\newblock In \emph{Advances in Neural Information Processing Systems}, 2021.

\bibitem[Touvron et~al.(2021)Touvron, Cord, Douze, Massa, Sablayrolles, and
  J{\'e}gou]{touvron2021training}
Hugo Touvron, Matthieu Cord, Matthijs Douze, Francisco Massa, Alexandre
  Sablayrolles, and Herv{\'e} J{\'e}gou.
\newblock Training data-efficient image transformers \& distillation through
  attention.
\newblock In \emph{International Conference on Machine Learning}, 2021.

\bibitem[Wang et~al.(2020)Wang, Wohlwend, and Lei]{wang2020structured}
Ziheng Wang, Jeremy Wohlwend, and Tao Lei.
\newblock Structured pruning of large language models.
\newblock In \emph{Conference on Empirical Methods in Natural Language
  Processing (EMNLP)}, 2020.

\bibitem[Xia et~al.(2022)Xia, Zhong, and Chen]{xia2022structured}
Mengzhou Xia, Zexuan Zhong, and Danqi Chen.
\newblock Structured pruning learns compact and accurate models.
\newblock \emph{arXiv preprint arXiv:2204.00408}, 2022.

\bibitem[Xiao(2009)]{xiao2009dual}
Lin Xiao.
\newblock Dual averaging method for regularized stochastic learning and online
  optimization.
\newblock \emph{Advances in Neural Information Processing Systems}, 2009.

\bibitem[Xie et~al.(2020)Xie, Dai, Chen, Dai, Zhao, Zha, Wei, and
  Pfister]{xie2020differentiable}
Yujia Xie, Hanjun Dai, Minshuo Chen, Bo~Dai, Tuo Zhao, Hongyuan Zha, Wei Wei,
  and Tomas Pfister.
\newblock Differentiable top-k with optimal transport.
\newblock In \emph{Advances in Neural Information Processing Systems}, 2020.

\bibitem[Yu et~al.(2012)Yu, Seide, Li, and Deng]{yu2012exploiting}
Dong Yu, Frank Seide, Gang Li, and Li~Deng.
\newblock Exploiting sparseness in deep neural networks for large vocabulary
  speech recognition.
\newblock In \emph{IEEE International Conference on Acoustics, Speech and
  Signal Processing (ICASSP)}, 2012.

\bibitem[Yun et~al.(2019)Yun, Han, Oh, Chun, Choe, and Yoo]{yun2019cutmix}
Sangdoo Yun, Dongyoon Han, Seong~Joon Oh, Sanghyuk Chun, Junsuk Choe, and
  Youngjoon Yoo.
\newblock Cutmix: Regularization strategy to train strong classifiers with
  localizable features.
\newblock In \emph{IEEE/CVF International Conference on Computer Vision}, 2019.

\bibitem[Zhang et~al.(2018)Zhang, Cisse, Dauphin, and
  Lopez-Paz]{zhang2018mixup}
Hongyi Zhang, Moustapha Cisse, Yann~N Dauphin, and David Lopez-Paz.
\newblock mixup: Beyond empirical risk minimization.
\newblock In \emph{International Conference on Learning Representations}, 2018.

\bibitem[Zhang et~al.(2022)Zhang, Lin, Chen, Xu, Chao, Shen, Li, Wu, and
  Ji]{zhang2022optimizing}
Yuxin Zhang, Mingbao Lin, Mengzhao Chen, Zihan Xu, Fei Chao, Yunhan Shen,
  Ke~Li, Yongjian Wu, and Rongrong Ji.
\newblock Optimizing gradient-driven criteria in network sparsity: Gradient is
  all you need.
\newblock \emph{arXiv preprint arXiv:2201.12826}, 2022.

\bibitem[Zhou et~al.(2021)Zhou, Zhang, Xu, and Zhang]{zhou2021effective}
Xiao Zhou, Weizhong Zhang, Hang Xu, and Tong Zhang.
\newblock Effective sparsification of neural networks with global sparsity
  constraint.
\newblock In \emph{Proceedings of the IEEE/CVF Conference on Computer Vision
  and Pattern Recognition}, 2021.

\bibitem[Zhu and Gupta(2018)]{zhu2018prune}
Michael Zhu and Suyog Gupta.
\newblock {To Prune, or Not to Prune: Exploring the Efficacy of Pruning for
  Model Compression}.
\newblock In \emph{International Conference on Learning Representations}, 2018.

\end{thebibliography}
\bibliographystyle{plainnat}

\newpage
\begin{appendix}

\section{Derivation of the Optimal Transport LP (Problem~\ref{eq:ot})}

Here, we show how the original top-$k$ LP with costs $c$ can be straightforwardly rewritten in the form of an optimal transportation problem.
For a given value vector $v \in \mathbb{R}^d$, cost vector $c \in \mathbb{R}_{++}^d$, and budget $k$, the top-$k$ LP is:
\begin{align*}
    \underset{m\in\mathbb{R}^d}{\rm maximize}\quad & v^T m \\
    \text{subject to}\quad & 0 \preceq m \preceq \one_d, \\
                           & c^T m = k.
\end{align*}
Define $y \coloneqq c \odot m$ and substitute to obtain:
\begin{align*}
    \underset{y\in\mathbb{R}^d}{\rm maximize}\quad & (v/c)^T y \\
    \text{subject to}\quad & 0 \preceq y \preceq c,  \\
                      & \one_d^T y = k.
\end{align*}
Now eliminate the upper bound constraint by introducing additional variables $y'$ to give:
\begin{align*}
    \underset{y, y' \in \mathbb{R}^d}{\rm maximize}\quad & (v/c)^T y \\
    \text{subject to}\quad &y \succeq 0, \; y' \succeq 0, \\
        & y + y' = c, \nonumber \\
        & \one_d^T y = k. \nonumber
\end{align*}
Finally, we rewrite this problem in the form of a standard OT problem by introducing the variable $Y \coloneqq [y,\,y']$ and the cost matrix $C \coloneqq [-v/c, \, 0]$ to yield:
\begin{align*}
    \underset{Y \in \mathbb{R}_+^{d \times 2}}{\rm minimize}\quad & \sum_{ij} C_{ij} Y_{ij}  \\
    \text{subject to}\quad  &Y \one_2 = c, \\ &\one_d Y = [k,\, \one^T_d c - k], 
\end{align*}
where we obtain the second column constraint by combining $y + y' = c$ and $\one_d^T y = k$.

\section{Derivation of Soft Top-$k$ Forward Pass (Algorithm \ref{alg:softtopk-forward})}
\label{sec:forward-derivation}

Problem~\ref{eq:reg-ot} is an entropy regularized optimal transport problem with cost matrix $C\coloneqq [-v/c, 0] \in \mathbb{R}^{d\times 2}$, row marginals $c \in \mathbb{R}^d$, and column marginals $[k, \one_d^T c - k]$. 
By Lemma 2 in \citep{cuturi2013sinkhorn}, the optimal solution to this problem can be written in the form $\mathrm{diag}(\exp \nu) \exp(-\beta C) \mathrm{diag}(\exp [\mu, \mu'])$,
with dual variables $\mu, \mu' \in \mathbb{R}$ and $\nu \in \mathbb{R}^d$. Moreover, we can compute a sequence of iterates converging to an optimal collection of dual variables using Sinkhorn iteration.

Note that the dual variables $\mu, \mu',$ and $\nu$ are only unique up to an additive constant: for any $\mu, \mu', \nu,$ and $\delta$, we have $\mathrm{diag}(\exp \nu) \exp(-\beta C) \mathrm{diag}(\exp [\mu, \mu']) = \mathrm{diag}(\exp (\nu + \delta)) \exp(-\beta C) \mathrm{diag}(\exp [\mu - \delta, \mu' - \delta])$.
We use this degree of freedom to fix $\mu' = 0$.

This yields the following Sinkhorn updates:
\begin{align*}
    \mu_{t+1} &= \log k - \log \sum_{i=1}^d \exp(-\beta C_{i1} + \nu_{t,i}) \\&= \log k - \log \sum_{i=1}^d \exp(\beta v_i / c_i + \nu_{t,i}),\\
    \nu_{t+1} &= \log c - \log(\exp(-\beta C_{\cdot 1} + \mu_t) + \exp(-\beta C_{\cdot 2} + \mu_t')) \\&= \log c - \log(\exp(\beta v / c + \mu_t) + \one_{d}),
\end{align*}
as stated in Algorithm~\ref{alg:softtopk-forward}.
Note that eliminating $\mu'$ reduces the cost of computing the forward pass by roughly $1/3$.

\section{Derivation of Soft Top-$k$ Backward Pass (Algorithm \ref{alg:softtopk-backward})}
\label{sec:backward-derivation}

The gradient of the loss with respect to the input values $v$ of the soft top-$k$ function follows from Theorem 3 of \citep{xie2020differentiable}
with some algebraic manipulation.
We restate the theorem below with the notation used in this paper.
In the following, let $\bar{q} \coloneqq q_{:{-1}}$ denote the vector $q$ with the last entry removed, and let $\bar{Y} \coloneqq Y_{\cdot,:-1}$ denote the matrix $Y$ with the last column removed.

\noindent\fbox{
    \parbox{\linewidth}{
\minihead{Theorem \citep{xie2020differentiable}}
Let the solution $Y \in \mathbb{R}^{N\times M}$ of the entropy regularized optimal transport problem with 
cost matrix $C$, row marginals $p$ and column marginals $q$ be given by:
\begin{equation*}
    Y_{ij} = \exp(-\beta C_{ij} + \nu^*_{i} + \mu^*_{j}),
\end{equation*}
where $\nu^* \in \mathbb{R}^N$, and $\mu^* \in \mathbb{R}^M$ are optimal dual variables.

Then $\frac{d\nu^*}{dC}$ and $\frac{d\mu^*}{dC}$ are given by:
\begin{equation*}
\begin{bmatrix}
    \frac{d\nu^*}{dC} \\
    \frac{d\mu^*}{dC}
\end{bmatrix}
=
\begin{bmatrix}
    -H^{-1}D \\
    \mathbf{0}
\end{bmatrix},
\end{equation*}
where $-H^{-1}D \in \mathbb{R}^{(N+M-1) \times N \times M}$, $\mathbf{0} \in \mathbb{R}^{1 \times N \times M},$ and
\begin{align*}
    D_{\ell ij} &= \beta \begin{cases} \delta_{\ell i} Y_{ij}, \quad \ell = 1, \dots, N  \\ \delta_{(\ell - N)j} Y_{ij}, \quad \ell = N+1, \dots, N+M-1 \end{cases},\\
    H^{-1} &= -\begin{bmatrix} (\mathrm{diag}(p))^{-1} + (\mathrm{diag}(p))^{-1} \bar{Y} \mathcal{K}^{-1} \bar{Y}^T (\mathrm{diag}(p))^{-1} & -(\mathrm{diag}(p))^{-1} \bar{Y} \mathcal{K}^{-1} \\ -\mathcal{K}^{-1} \bar{Y}^T (\mathrm{diag}(p))^{-1} & \mathcal{K}^{-1} \end{bmatrix},\\
    \mathcal{K} &= \mathrm{diag}(\bar{q}) - \bar{Y}^T (\mathrm{diag}(p))^{-1} \bar{Y}.
\end{align*}
    } 
} 

\vspace{0.5em}

Specializing this theorem to our setting with $N = d$ and $M = 2$, and recalling that the soft top-$k$ mask values are given by $m_i = Y_{i1} / c_i$, we obtain:
\begin{align*}
    H^{-1} &= \frac{1}{\mathcal{K}} \begin{bmatrix} -\mathcal{K}\mathrm{diag}(1 / c) - m m^T &  m \\  m^T  & -1 \end{bmatrix}, \\
    \mathcal{K} &= k - \one^T (m^2 \odot c).
\end{align*}
Therefore:
\begin{equation*}
    \begin{bmatrix}
        \frac{d\nu^*}{dC_{\cdot1}} \\
        \frac{d\mu^*}{dC_{\cdot1}}
    \end{bmatrix}
    = 
    \frac{\beta}{\mathcal{K}} \begin{bmatrix} 
        \mathcal{K}\mathrm{diag}(m) + m (m^2 \odot c)^T -  m (m\odot c)^T \\
        -(m^2 \odot c)^T + (m \odot c)^T
    \end{bmatrix}.
\end{equation*}
Taking the derivative of the mask with respect to the first column of the cost matrix, we have:
\begin{align*}
    \frac{dm}{dC_{\cdot 1}} &= -\beta \mathrm{diag}(m) + m\frac{d\mu^*}{dC_{\cdot 1}} + \mathrm{diag}(m) \frac{d\nu^*}{dC_{\cdot 1}} \\
     &= -\beta \left(\mathrm{diag}(m \odot (\one - m)) - \frac{1}{\mathcal{K}} (m \odot (\one - m)) (m \odot (\one - m) \odot c)^T\right).
\end{align*}
Finally, using our definition of the cost matrix as $C_{\cdot 1} = -v / c$ and applying the loss gradient $g \coloneqq \left(\frac{dL}{dm}\right)^T$, we obtain:
\begin{align*}
    \left(\frac{dL}{dv}\right)^T = \left(\frac{dL}{dm} \frac{dm}{dv}\right)^T = \beta m \odot (\one - m) \odot \left(\frac{g}{c} - \frac{1}{\mathcal{K}} \one^T(g \odot m \odot (\one - m)) \right),
\end{align*}
as desired.

\section{Training Details}

In Tables~\ref{tab:resnet50-hyperparams} and \ref{tab:vit-hyperparams}, we detail the hyperparameters used for our training runs.
These hyperparameters were derived from \url{https://github.com/pytorch/vision/blob/96d1fecf/references/classification/README.md}.

\begin{table}[H]
    \caption{ResNet-50 training hyperparameters}
    \label{tab:resnet50-hyperparams}
    \centering
    \begin{tabular}{rl}
        \toprule
        \textbf{Hyperparameter} & \textbf{Value} \\
        \midrule
        optimizer & Nesterov accelerated gradient method ($\mathrm{momentum} = 0.9$) \\
        max. learning rate & $1.0$ \\
        min. learning rate & $0.001$ \\
        learning rate warmup epochs & $5$ \\
        learning rate decay schedule & cosine \\
        batch size & $4096$ \\
        weight decay & $10^{-4}$ ($0.0$ for bias and normalization parameters) \\
        label smoothing & $0.1$ \\
        data augmentation & random crops, random horizontal flips\\
        input resolution & $224\times 224$ \\
        sparsity annealing schedule & linear from $1$ to target sparsity at epoch fraction $0.2$ \\
        $\beta$ annealing schedule & linear from $1$ to $\beta_\mathrm{max}$ at epoch fraction $0.8$ \\
        Sinkhorn max. iterations & $100$ \\
        Sinkhorn tolerance $\epsilon$ & $0.01$ \\
        \bottomrule
    \end{tabular}
\end{table}

\begin{table}[H]
    \caption{ViT-B/16 training hyperparameters}
    \label{tab:vit-hyperparams}
    \centering
    \begin{tabular}{rl}
        \toprule
        \textbf{Hyperparameter} & \textbf{Value} \\
        \midrule
        optimizer & AdamW ($\beta = (0.9, 0.999), \epsilon = 10^{-8}$)\\
        max. learning rate & $0.003$ \\
        min. learning rate & $0.0$ \\
        learning rate warmup epochs & $10\%$ of total epochs \\
        learning rate decay schedule & cosine \\
        batch size & $4096$ \\
        weight decay & $0.3$ ($0.0$ for bias and normalization parameters) \\
        label smoothing & $0.1$ \\
        data augmentation & random crops, random horizontal flips, \\ & RandAugment (ops $=2$, magnitude $=9$) \\
        mixup $\alpha$ & $0.2$ \\
        CutMix $\alpha$ & $1.0$ \\
        gradient $L_2$ norm clip & $1.0$ \\
        input resolution & $224 \times 224$ \\
        exponential moving averaging & false \\
        sparsity annealing schedule & linear from $1$ to target sparsity at epoch fraction $0.2$ \\
        $\beta$ annealing schedule & linear from $1$ to $\beta_\mathrm{max}$ at epoch fraction $0.8$ \\
        Sinkhorn max. iterations & $100$ \\
        Sinkhorn tolerance $\epsilon$ & $0.01$ \\
        \bottomrule
    \end{tabular}
\end{table}

\section{FLOP Measurements}
\label{sec:flop-measurements}

Due to differences in the computational cost associated with individual parameters, the sparsity fraction does not map 1-to-1 to
the fraction of FLOPs required for inference. Tables~\ref{table:resnet50-flops} and \ref{table:vit-flops} give FLOP costs for our
sparse models as a percentage of the FLOP cost of the corresponding dense model. We performed our FLOP measurements using the
open source tool available at \url{https://github.com/sovrasov/flops-counter.pytorch/}.

We count multiply and add operations as one FLOP each. We remark that there exists some inconsistency in the literature regarding this convention, 
with some prior work using multiply-accumulate (MAC) counts and FLOP counts interchangeably. 
To convert the base FLOP counts listed below for ResNet-50 and ViT-B/16 to MACs, we can simply divide the given counts by 2.

In Table~\ref{table:resnet50-flops}, the ResNet-50 FLOP counts for Top-KAST are slightly higher than those for Spartan,
possibly due to the exclusion of the input convolutional layer from pruning in the case of Top-KAST.\footnote{This setup for Top-KAST follows the protocol used by \citet{jayakumar2020top}}
In particular, this demonstrates that the accuracy improvements seen in Spartan-trained models over those trained using Top-KAST 
do not correlate with an increase in their FLOP costs.

\begin{table}[H]
    \caption{Percentage FLOP costs of sparse ResNet-50 models relative to the FLOP cost of a dense ResNet-50.
        The cost of a dense ResNet-50 model is 8.24 GFLOPs.}
    \label{table:resnet50-flops}
    \centering
    \begin{tabular}{rcccccc}
        \toprule
               &        & \multicolumn{5}{c}{\textbf{Sparsity}} \\
               \cmidrule{3-7}
        \textbf{Method} & \textbf{Epochs} & \textbf{80\%} & \textbf{90\%} & \textbf{95\%} & \textbf{97.5\%} & \textbf{99\%} \\
        \midrule
        \multirow{3}{*}{Top-KAST} & $100$ & \res{28.43}{0.28} & \res{17.11}{0.88} & \res{11.21}{0.29} & \res{7.82}{0.08} & - \\
                                  & $200$ & \res{27.80}{0.31} & \res{17.66}{0.07} & \res{11.75}{0.12} & \res{7.93}{0.13} & - \\
                                  & $400$ & \res{28.06}{0.12} & \res{18.05}{0.28} & \res{11.40}{0.16} & \res{7.74}{0.19} & - \\
                 \cmidrule(lr){3-7}
        \multirow{3}{*}{Spartan}  & $100$ & \res{24.68}{0.07} & \res{14.48}{0.11} & \res{8.67}{0.14} & \res{5.04}{0.11} & \res{2.53}{0.04} \\
                                  & $200$ & \res{24.37}{0.60} & \res{14.37}{0.09} & \res{8.43}{0.06} & \res{4.97}{0.07} & \res{2.59}{0.04} \\
                                  & $400$ & \res{23.97}{0.31} & \res{14.20}{0.08} & \res{8.56}{0.10} & \res{5.07}{0.06} & \res{2.50}{0.03} \\
        \bottomrule
    \end{tabular}
\end{table}

\begin{table}[H]
    \caption{Percentage FLOP costs of 90\% sparse ViT-B/16 models at $224\times 224$ input resolution relative to the FLOP cost of a dense ViT-B/16 model.
    The cost of a dense ViT-B/16 model is 35.19 GFLOPs.}
    \label{table:vit-flops}
    \centering
    \begin{tabular}{rcccc}
        \toprule
          & & \multicolumn{3}{c}{\textbf{Sparsity Structure}}\\
          \cmidrule{3-5}
        \textbf{Method} & \textbf{Epochs} & \textbf{Unstructured} & $\mathbf{16\times 16}$ \textbf{blocks} & $\mathbf{32\times 32}$ \textbf{blocks} \\
        \midrule
        \multirow{2}{*}{Top-KAST} & $100$ & \res{13.42}{0.00} & \res{12.76}{0.00} & \res{12.76}{0.00} \\
                                  & $300$ & \res{13.41}{0.00} & \res{12.76}{0.00} & \res{12.76}{0.00} \\
          \cmidrule(lr){3-5}
        \multirow{2}{*}{Spartan}  & $100$ & \res{13.40}{0.00} & \res{12.76}{0.00} & \res{12.76}{0.00} \\
                                  & $300$ & \res{13.43}{0.00} & \res{12.76}{0.00} & \res{12.76}{0.00} \\
        \bottomrule
    \end{tabular}
\end{table}

\section{FLOP-Sensitive Pruning}
\label{sec:flop-sensitive-pruning}

We demonstrate FLOP-sensitive pruning with Spartan on ResNet-50 using the following cost model: assign a cost of 1 to each parameter of
a fully connected layer, and a cost of $N^2$ to each parameter of a convolutional layer where the output has size $N \times N$ along its spatial dimensions.
We evaluate two valuation functions: $v_1(c_i, \theta_i) = c_i |\theta_i|$ and $v_{0.5}(c_i, \theta_i) = \sqrt{c_i} |\theta_i|$.
$v_1$ results in the same pruning order as in standard pruning, but with a FLOP budget constraint instead of the usual sparsity budget.
$v_{0.5}$ assigns a lower value to the parameters of convolutional layers, and results in networks where the parameters of convolutional layers are preferentially pruned.
We use $\beta_\mathrm{max} = 10$ for $v_1$ and $\beta_\mathrm{max} = 160$ for $v_{0.5}$ to compensate for the relatively smaller scale of the
normalized values $v/c$ in the soft top-$k$ forward pass (Algorithm~\ref{alg:softtopk-forward}).

Table~\ref{table:flop-pruning} gives the top-1 accuracy, FLOP percentage, and sparsity percentage for each of these valuation functions.
Spartan yields models with identical FLOP percentages of $5.76\%$, which is slightly higher than the budgeted value of $5\%$---this discrepancy is due to the additional cost of the normalization layers and activation functions in the network.
Most notably, there is a substantial difference in the sparsity percentages realized by these valuation functions.
As expected, $v_{0.5}$ preferentially sparsifies the parameters of convolutional layers and yields denser fully connected layers, resulting in lower sparsity overall.

\begin{table}[H]
    \caption{Results of FLOP-sensitive pruning experiments on ImageNet-1K with ResNet-50 models.}
    \label{table:flop-pruning}
    \centering
    \begin{tabular}{rccc}
    \toprule
        & \textbf{Accuracy \%} & \textbf{FLOP \%} & \textbf{Sparsity \%} \\
    \midrule
        $v_1$     & $73.88$ & $5.76$ & $95.94$ \\
        $v_{0.5}$ & $74.13$ & $5.76$ & $89.73$ \\
    \bottomrule
    \end{tabular}
\end{table}

\section{Additional Experiments}

In Table~\ref{table:additional-ablations}, we compare Spartan against two additional variants of the Top-KAST baseline:
Top-KAST with the Erdos-Renyi-Kernel (ERK) sparsity distribution, and with pruning applied to the parameters of all convolutional and fully-connected layers with the exception of bias terms (prune all).
Top-KAST (excl. input conv.) denotes the Top-KAST variant used in the experiments presented in the main text, where we exclude the input convolutional layer from pruning.
We find that there is some small variation in the measured top-1 validation accuracies, but our conclusion that Spartan improves generalization at higher levels of sparsity is unchanged.

\begin{table}[H]
    \caption{Comparison between Spartan and additional variants of the Top-KAST baseline on ImageNet-1K with ResNet-50 models.}
    \label{table:additional-ablations}
    \centering
    \begin{tabular}{rcccc}
        \toprule
               &        & \multicolumn{2}{c}{\textbf{Sparsity}} \\
               \cmidrule{3-4}
        \textbf{Method} & \textbf{Epochs} & \textbf{90\%} & \textbf{95\%} \\
        \midrule
        \multirow{3}{*}{Top-KAST (ERK)} & $100$ & $75.98$ & $73.63$ \\
                                  & $200$ & $77.06$ & $75.08$ \\
                                  & $400$ & $77.48$ & $75.62$ \\
                 \cmidrule(lr){3-4}
        \multirow{3}{*}{Top-KAST (prune all)} & $100$ & $75.74$ & $73.72$ \\
                                    & $200$ & $76.77$ & $75.07$ \\
                                    & $400$ & $77.44$ & $75.77$ \\
                \cmidrule(lr){3-4}
        \multirow{3}{*}{Top-KAST (excl. input conv.)} & $100$ & \res{75.48}{0.15} & \res{73.51}{0.16} \\
                                    & $200$ & \res{76.84}{0.11} & \res{75.20}{0.11} \\
                                    & $400$ & \res{77.37}{0.07} & \res{75.90}{0.04} \\
                \cmidrule(lr){3-4}
        \multirow{3}{*}{Spartan}  & $100$ & \res{76.17}{0.10} & \res{74.68}{0.24}  \\
                                  & $200$ & \res{77.06}{0.13} & \res{75.92}{0.01}  \\
                                  & $400$ & \res{77.40}{0.06} & \res{76.48}{0.20}  \\
                                  \bottomrule
    \end{tabular}
\end{table}

\section{Learned Sparsity Patterns}

\begin{figure}[H]
    \centering
    \includegraphics[width=\textwidth]{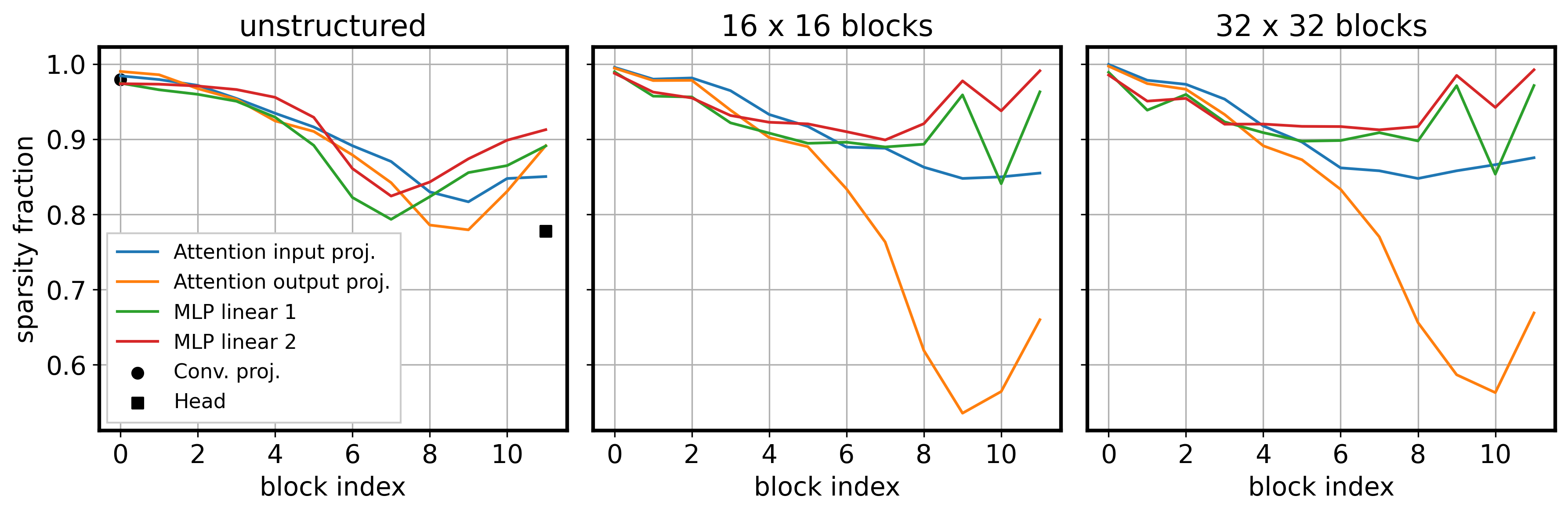}
    \caption{Per-layer sparsities of ViT-B/16 models trained with Spartan.}
    \label{fig:vit-sparsity}
\end{figure}

We observe a qualitative difference in the distribution of per-layer sparsities between ViT-B/16 models trained with unstructured sparsity and 
those trained with block structured sparsity (Figure~\ref{fig:vit-sparsity}).
In particular, the output projections of self-attention layers under block structured pruning are significantly more dense 
in the later blocks of the network relative to unstructured pruning. 
The reasons for this difference are not immediately clear to us, and we leave further investigation of this phenomenon to future work.

\begin{figure}[H]
    \centering
    \includegraphics[width=\textwidth]{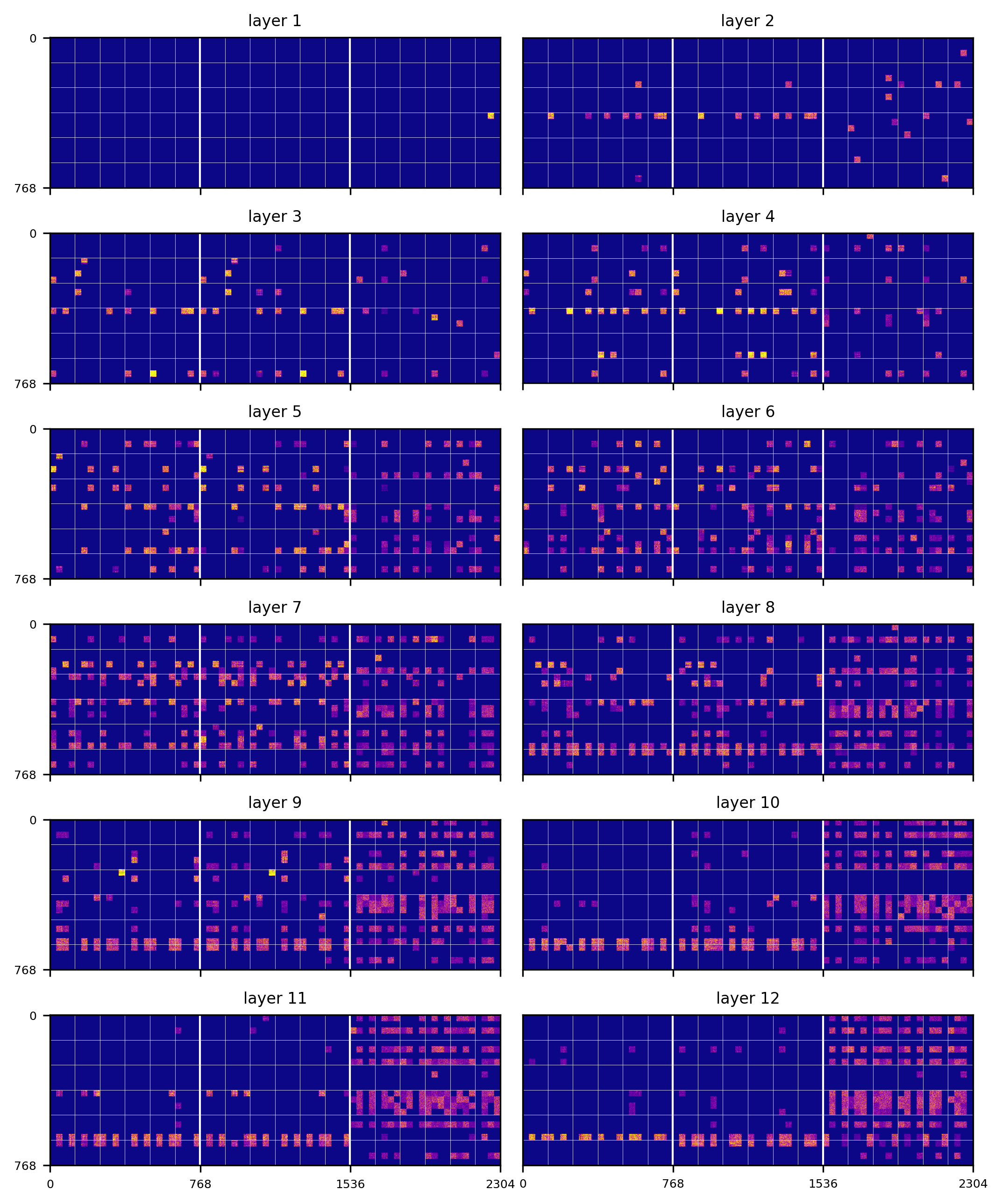}
    \caption{Weight magnitudes of the input projection matrices of each self-attention layer in a $32\times 32$ block sparse ViT-B/16 network trained using Spartan.}
    \label{fig:vit-sparsity-inproj}
\end{figure}

Block structured pruning produces coherent sparsity patterns in ViT-B/16 models.
In Figure~\ref{fig:vit-sparsity-inproj}, we visualize the magnitudes of the weight matrices corresponding to the input
projection of each self-attention layer in a ViT-B/16 model trained with Spartan using $32\times 32$ block structured pruning.
This matrix maps vectors of dimension $768$ to query, key, and value embedding vectors, each of dimension $768$.
We observe that the training process yields similar sparsity patterns in the query and key embedding submatrices, which 
correspond to the left and center panels in the visualization for each layer.
This is an intuitively reasonable property since the self-attention layer computes inner products of the query and key embeddings
in order to construct attention maps.
We note that this symmetry emerges purely as a result of the optimization process; we did not incorporate any prior knowledge into Spartan
regarding the role of particular entries of the weight matrices subject to sparsification.

\end{appendix}

\end{document}